\documentclass[a4paper]{article}

\title{Human Perception of Intrinsically Motivated Autonomy in Human-Robot 
Interaction}
\author{Marcus M. Scheunemann$^1$\footnote{Contact corresponding author 
using 
    \href{mailto:marcus@mms.ai}{marcus@mms.ai} or visit 
    \url{https://mms.ai}.} \and 
  Christoph Salge$^1$ \and Daniel Polani$^1$ \and Kerstin Dautenhahn$^{1,2}$
}
\date{%
  {\small{$^{1}$Adaptive Systems Research Group, University of 
  Hertfordshire, 
      Hatfield, United Kingdom\\%
      $^{2}$Social and Intelligent Robotics Research Laboratory,
      University of Waterloo, Waterloo, Canada}}\\[2ex]%
  \today
}

\usepackage[svgnames]{xcolor}
\usepackage[hyperindex,breaklinks]{hyperref}
\hypersetup{
  colorlinks  = true,
  urlcolor    = blue, % external hyperlinks
  linkcolor   = blue,
  citecolor   = Crimson,  
  pdfinfo     = {
    Title    = {Human Perception of Intrinsically Motivated Autonomy in 
    Human-Robot Interaction},
    Author   = {Marcus M. Scheunemann, Christoph Salge, Daniel Polani, Kerstin Dautenhahn},
    Subject  = {A challenge in using robots in human-inhabited environments is
      to design behavior that is engaging, yet robust to the perturbations 
      induced by human interaction. Our idea is to imbue the robot with 
      intrinsic motivation (IM) so that it can handle new situations and 
      appears as a genuine social other to humans and thus be of more interest 
      to a human interaction partner.
      Human-robot interaction (HRI) experiments mainly focus on scripted or 
      teleoperated robots, that mimic characteristics such as IM to control 
      isolated behavior factors. This article presents a "robotologist" study 
      design that allows comparing autonomously generated behaviors with each 
      other, and, for the first time, evaluates the human perception of 
      IM-based generated behavior in robots.
      We conducted a within-subjects user study (N=24) where participants 
      interacted with a fully autonomous Sphero BB8 robot with different 
      behavioral regimes: one realizing an adaptive, intrinsically motivated 
      behavior and the other being reactive, but not adaptive.
      The robot and its behaviors are intentionally kept minimal to concentrate 
      on the effect induced by IM.
      A quantitative analysis of post-interaction questionnaires showed a 
      significantly higher perception of the dimension "Warmth" compared to 
      the reactive baseline behavior. Warmth is considered a primary dimension 
      for social attitude formation in human social cognition. A human 
      perceived as warm (friendly, trustworthy) experiences more positive 
      social interactions.},
    Keywords = {Intrinsic Motivation, User Study, Human-Robot Interaction, 
    Social Cognition, Predictive Information, Embodied Cognition},
    Lang     = {en-US}
  }
}

\usepackage[verbose=true,a4paper]{geometry}

\usepackage{booktabs,csvsimple,bigdelim}
\usepackage[
binary-units=true,
range-units = single, %e.g., single: 1 to 2 m, repeat: 1 m to 2 m
detect-all]{siunitx}

\usepackage{pdftexcmds}
\makeatletter
\newcommand*{\COND}[1]{%
  \stringcases
  {#1}%
  {%
    {balanced-REA}{\ensuremath{REA}}%
    {balanced-ADA}{\ensuremath{ADA_b}}%      
    {ADA}{\ensuremath{ADA}}%
  }%
  {[nada]}%
}
\newcommand{\stringcases}[3]{%
  \romannumeral
  \str@case{#1}#2{#1}{#3}\q@stop
}
\newcommand{\str@case}[3]{%
  \ifnum\pdf@strcmp{\unexpanded{#1}}{\unexpanded{#2}}=\z@
  \expandafter\@firstoftwo
  \else
  \expandafter\@secondoftwo
  \fi
  {\str@case@end{#3}}
  {\str@case{#1}}%
}
\newcommand{\str@case@end}{}
\long\def\str@case@end#1#2\q@stop{\z@#1}
\makeatother
\usepackage{glossaries}
\glsdisablehyper % there won't be an actual glossary?
\newacronym{PI}{PI}{predictive information}
\newacronym{TiPI}{TiPI}{time-local predictive information}
\newacronym{IM}{IM}{intrinsic motivation}
\newacronym{HRI}{HRI}{human-robot interaction}
\newacronym{IMU}{IMU}{inertial measurement unit}
\newacronym{RoSAS}{RoSAS}{Robotic Social Attributes Scale}
\newacronym{SDT}{SDT}{Self-Determination Theory}
\newacronym{BLE}{BLE}{Bluetooth Low Energy}

\usepackage[font={small,it}]{caption,subcaption}
\usepackage[onehalfspacing]{setspace}
\usepackage{amsmath,amssymb, amstext}
\usepackage{graphicx}

\usepackage{apacite}
\usepackage{natbib}
\newcommand*{\doi}{}
\makeatletter
\newcommand{\doi@}[1]{\href{https://doi.org/#1}{#1}}
\DeclareRobustCommand{\doi}{\hyper@normalise\doi@}
\makeatother

\usepackage{booktabs,bigdelim}
\usepackage[
binary-units=true,
range-units = single, %e.g., single: 1 to 2 m, repeat: 1 m to 2 m
detect-all]{siunitx}

\def\keywordname{{\bfseries Keywords}}%
\def\keywords#1{\par\addvspace\medskipamount{\rightskip=0pt plus1cm
    \def\and{\ifhmode\unskip\nobreak\fi\ $\cdot$
    }\noindent\keywordname\enspace\ignorespaces#1\par}}

\usepackage{fancyhdr}
\newcommand{\headerleft}{\footnotesize Human Perception of Intrinsically 
Motivated Autonomy in HRI}
\newcommand{\headerright}{\footnotesize preprint}
\fancyheadoffset{0pt}

\newcommand{\Autoref}[1]{%
  \begingroup%
  \def\chapterautorefname{Chapter}%
  \def\sectionautorefname{Section}%
  \def\subsectionautorefname{Subsection}%
  \autoref{#1}%
  \endgroup%
}

\begin{document}

\maketitle

\begin{abstract}
A challenge in using robots in human-inhabited environments is to design 
behavior that is engaging, yet robust to the perturbations induced by human 
interaction. Our idea is to imbue the robot with intrinsic motivation~(IM) so 
that it can handle new situations and appears as a genuine social other to 
humans and thus be of more interest to a human interaction partner.
Human-robot interaction~(HRI) experiments mainly focus on scripted or 
teleoperated robots, that mimic characteristics such as IM to control isolated 
behavior factors. This article presents a ``robotologist'' study design that 
allows comparing autonomously generated behaviors with each other, and, for the 
first time, evaluates the human perception of IM-based generated behavior in 
robots.
We conducted a within-subjects user study~($N=24$) where participants 
interacted with a fully autonomous Sphero BB8 robot with different behavioral 
regimes: one realizing an adaptive, intrinsically motivated behavior and the 
other being reactive, but not adaptive.
The robot and its behaviors are intentionally kept minimal to concentrate on 
the effect induced by IM.
A quantitative analysis of post-interaction questionnaires showed a 
significantly higher perception of the dimension ``Warmth'' compared to the 
reactive baseline behavior. Warmth is considered a primary dimension for social 
attitude formation in human social cognition. A human perceived as 
warm~(friendly, trustworthy) experiences more positive social interactions.
\end{abstract}


\maketitle

\glsresetall %
\section{Introduction}
This article is part of a larger research program to produce autonomous robots, 
i.e., robots that are not teleoperated or remotely controlled, yet robust to 
the unknown perturbations and capable of sustained interaction with humans.
In this study, we look at the effects that an intrinsically motivated robot 
behavior has on the social perceptions of robots by humans, and whether they 
may engage human participants in an interaction.
An example of an intrinsically motivated behavior is a child interacting with 
a puppy. The child will likely be motivated to play with a puppy, even without 
an external reward (such as promised money) and even without the existence of 
an extrinsic reward (such as playing with the puppy as a means to an end, i.e., 
to train it).
Instead, the motivation for the interaction might result purely from wanting 
to do this activity for its own sake, i.e., the child is intrinsically 
motivated to play with the puppy.
Our long term goal with this research is to better understand how to engage the
human, similar to the child in the example.
But here we do not focus on modeling the intrinsic motivation of the human
interaction partner, but rather focus on using a computational model of 
intrinsic motivation to generate robot behavior. 
Our idea is that a robot that is intrinsically motivated (driven by a specific, 
adaptive IM model) is perceived more like a social other, and thus is more 
engaging for a human interaction partner.

There are approaches to keep humans engaged in the interaction with a robot. 
For example, \citet{PinillosMarcosEtAl-16} developed an 
autonomous hotel robot. 
It attracts the attention of the hotel guests, many of them wanting to know 
more about the robot itself.
They propose that the robot's services (i.e., its competence or usefulness) 
need to be large in order to keep customers engaged. 
On the other hand, \citet{KandaShiomiEtAl-10}
developed a semi-teleoperated mall robot and incrementally added novel 
behaviors, such as self-disclosure. A field trial indicates that the robot 
attracted reoccurring visitors, without increasing its services.
Engagement is also a concern in the field of social robotics in 
education~\citep{BelpaemeKennedyEtAl-18}.
One existing approach here is to develop robots with a set of hand-designed 
questions, comments, and 
statements~\citep{GordonBreazealEtAl-15,CehaChhibberEtAl-19}. This makes 
the 
robots \emph{appear} curious, which elicits curiosity in the humans too, which 
in turn enhances learning and memory 
retention~\citep{OudeyerGottliebEtAl-16}.
Curiosity is part of the broader concept of intrinsic 
motivation~\citep{OudeyerGottliebEtAl-16}, or is even used synonymously 
with intrinsic motivation~\citep{Schmidhuber-91}.

The previous studies either constrained the context and focused on a specific 
task~\citep[e.g.,][]{GordonBreazealEtAl-15,PinillosMarcosEtAl-16}, or were 
relying on humans teleoperating the 
robot~\citep[e.g.,][]{KandaShiomiEtAl-10,CehaChhibberEtAl-19}.
Teleoperation, or the Wizard-of-Oz model, remains the state of the art for 
many HRI studies~\citep{ClabaughMataric-19}. This is due to the challenge 
to define a sufficient set of execution rules (i.e., behaviors) for an HRI 
task; this holds true even in a laboratory setting. It remains elusive to 
achieve autonomous, social behavior in an unconstrained environment,
i.e., for any given task or goal in the real 
world~\citep{ChristensenOkamuraEtAl-16,BelpaemeKennedyEtAl-18}.
Developing a robot driven by an actual intrinsic motivation formalism, such 
as the drive to explore its environment and its capabilities, might offer a 
solution. If successful, this would provide us with a robust behavior 
generation mechanism that allows us to ``Escape 
Oz''~\citep{ClabaughMataric-19}, while also producing behavior that appears 
curious, or similarly engaging to the human interaction partner. This will 
reduce the reliance on human adaptation or teleoperation, and could provide 
a promising pathway towards having robots more easily deployed in everyday 
life.
 
Our idea is that imbuing a robot with a computational model of \gls{IM} 
makes the robot appear as a genuine \emph{social other} -- similar maybe to 
an animal -- and thus be of more interest to a human interaction partner. 
The concept of intrinsic motivation originates in psychology, initially in 
close relation to \gls{SDT}~\citep{ryan2000self}. \gls{SDT} posits that 
humans have an inherent tendency to seek out novelty and challenges, to 
extend and exercise their capacities to explore and to learn, without 
having to be coerced by an extrinsic reward. According to \gls{SDT}, humans 
have inherent drives for competence, autonomy, and relatedness. 
Computational models of intrinsic motivation aim to formalize the 
principles that create those drives to make them operational, i.e., they 
can be used to create spontaneous exploration and curiosity in an 
artificial agent~\citep{OudeyerKaplan-09}. 
Therefore, we hypothesize that they will give an artificial agent a 
stronger social presence, and thus make them a more interesting interaction 
partner. 

The known models of intrinsic motivation have a range of interesting 
properties. The idea of universality is of particular interest for this 
application, in particular the fact that \glspl{IM} can cope with changes to
an agent's environment or its morphology~(we discuss this in 
more detail in \autoref{sec:IM}, \ref{sec:PI}, and \ref{sec:TiPI-summary}).
This makes this approach, in principle, suitable to be deployed on any robot
and it also allows it to deal with any environment or context. The biggest
limitation here is usually computational complexity.
The method is also limited by the fact that several approaches at least
require agent-centric forward models, similar to sensorimotor 
contingencies~\citep{o2001sensorimotor}, which might not be easily 
obtainable. 
Finally, most \glspl{IM} can be expressed to operate on the immediate 
perception-action loop of the robot, allowing for tightly coupled or 
entrained behavior with both the environment or other actors. Both of these 
properties make \glspl{IM} an interesting family of approaches to deploy in 
autonomous \gls{HRI} robots, as there is a requirement for interactive 
feedback on a short feedback loop and for the ability to robustly deal with 
a range of situations. 
This is particularly relevant, as human social cognition is believed to 
heavily depend on interaction -- and thus any approach that aims to 
encourage interaction should be robust to the perturbations induced by 
those social, and possibly physical interactions. 

In the remainder of this article, we want to substantiate this main idea 
with an \gls{HRI} study involving 24 human participants. As a first step, 
we evaluate the human perception of robots with different behavior with the 
help of post-experiment questionnaires. We compare how the introduction of 
intrinsically motivated behavior affects human perception, and discuss how 
these factors can lead to the formation of different social attitudes. Our 
main focus in this article is on the \emph{Warmth} dimension.
Warmth and \emph{Competence} are considered the two main dimensions in 
describing almost all social attitudes in human social cognition, such as 
friendliness, empathy, admiration, envy, contempt and 
pity~\citep{FiskeCuddyEtAl-07,AbeleHaukeEtAl-16}.
Warmth is considered the primary dimension for social characterizing peers. 
This means, when characterizing other people, we firstly judge their intent 
(Warmth) before judging their capability (Competence) to enact their intent. 
Warmth is strongly linked to the measure of 
trust~\citep{FiskeCuddyEtAl-07,Fiske-18}. A person who is perceived as warm 
is also perceived as more trustworthy. For example, \citet{KulmsKopp-18} 
use it as an indicator of people's trust 
in computers.
Importantly, from human social cognition, it is known that human's who are 
perceived as warm experience more positive social interactions compared to 
their peers who are perceived as less warm~\citep{FiskeCuddyEtAl-07}.
Recent research in \gls{HRI} has shown that human participants prefer to 
interact again with the robot if they perceive its behavior as more 
warm~\citep{OliveiraArriagaEtAl-19,ScheunemannCuijpersEtAl-20}.
Consequently, in order to welcome robots in our everyday life, an 
understanding is needed for how to enable the perception of Warmth for 
robots.

We will see that the robot that continues adaptation based on its intrinsic 
motivation, i.e., the intrinsically motivated robot, generates behavior that 
participants rated as more warm compared to the baseline behavior.
This is a step towards the long-term goal of producing a robot capable of 
sustained interaction, as it suggests a method to induce a positive social 
attitude towards the robot in the human. 
Further studies are, of course, needed to see if this effect for Warmth 
transfers from human-human interaction to human-robot interaction.
We also still need to investigate if higher perceived Warmth for a robot 
actually leads to more sustained interaction. The interplay between 
personality 
and social relationships is still an ongoing -- and complex -- investigation 
for human-human interaction~\citep{GeukesBreilEtAl-19}.
Our expectation, which needs to be confirmed in future work, is that a 
robot which is perceived as warm~(friendly, trustworthy) is more likely to 
receive more positive interactions from people and will facilitate 
long-term interactions.
\subsection*{Overview}
First, \autoref{sec:background} outlines the background on intrinsic 
motivation, its computational approaches, and its relation to autonomy, 
insofar it relates to the present work. \Autoref{sec:TiPI-overview} then 
introduces time-local predictive information~(TiPI), the information 
theoretic formalism we use to implement intrinsic motivation in our 
studies. We outline the concrete approximations (and their assumptions) to 
compute PI. In particular, we highlight how to make this approach suitable 
for deployment on an actual robot and why it is a good candidate for our 
research questions.

\Autoref{sec:study_design} presents the materials and methods of our 
within-subjects study~($N=24$). The study consists of two conditions with 
the same robot platform: the intrinsically motivated behavior in one 
condition is generated using predictive information maximization, the 
behavior in the other condition is a reactive baseline behavior.
The focus is on the interplay between the robot and the human participant. 
We designed a study where the participants interact and observe the robots 
in order to understand their behavioral differences, but they are not given 
information about the robot's task and they cannot order the robot to do 
something. Instead, the robot and the human participants explore their 
behavior towards each other.
We call this ``robotologist'' study design (for details see 
\autoref{sec:studyII:robotology}).
The study design is motivated by a preliminary study by 
\citet{ScheunemannSalgeEtAl-19} which has been conducted and published 
prior to this work. We outline the learned lessons from this previous study 
where needed.

\Autoref{sec:studyII:results} presents the results, concentrating on our 
two main hypotheses: one focusing on the perceived Warmth of the 
intrinsically motivated robot behavior, and the other on the lack of 
difference in Perceived Intelligence and Competence between the robot 
scenarios.
We found that our study design makes both robot behaviors appear similarly 
competent. This is important in order to focus solely on the effect on the 
Warmth dimension, without interfering by the Competence rating.
Most importantly, the study provides evidence that the intrinsically 
motivated robot displays behavior that is perceived as more warm compared 
to the reactive baseline behavior.
\Autoref{sec:discussion} discusses the implications of those findings, and 
how they can be applied to other projects. \Autoref{sec:conclusion} 
summarizes the study and concludes the article.

\section{Background}
\label{sec:background}
This section provides some background of the previously mentioned concepts 
relating to this work. 

\subsection{Intrinsic Motivation}
\label{sec:IM}
\glsreset{IM}
A common definition of \gls{IM} in psychology is ``doing [\dots] an 
activity for its inherent satisfactions rather than for some separable 
consequence''~\citep{RyanDeci-00}. An intrinsically motivated agent is 
moved to do something, for the enjoyment of the activity itself, for the 
``fun or [the] challenge entailed rather than because of external products, 
pressures, or rewards''~\citep[pg.~56]{RyanDeci-00}.
Since intrinsic motivations have been considered an instrumental ingredient 
in the development of humans \citep{OudeyerKaplanEtAl-07}, there has also 
been a great interest in developmental robotics to produce formalized 
models that can be used to imbue robots with drives for competence and 
knowledge acquisition.
Although concepts like ``fun'' and ``challenge'' are presented as crucial 
for the definition of IM in psychology, the literature lacks consensus on 
what these concepts are~\citep{OudeyerKaplan-09}. This missing consensus 
and the resulting vagueness of the definition makes it impossible to 
transfer it directly onto a robotic system.
\citet{OudeyerKaplan-08} characterize intrinsic motivation in the 
following, broadly accepted way:
\begin{quote}
An activity or an experienced situation, be it physical or imaginary, is 
intrinsically motivating for an autonomous entity if its interest depends 
primarily on the collation or comparison of information from different 
stimuli [...]. [...] the information that is compared has to be understood 
in an information theoretic perspective [...], independently of their 
meaning. As a consequence, measures which pre-suppose the meaning of 
stimuli, i.e., the meaning of sensorimotor channels (e.g., the fact that a 
measure is a measure of energy or temperature or color), do not 
characterize intrinsically motivating activities or situations.
\end{quote}

Nowadays, there is a range of formal models that roughly fall under the 
header of intrinsic motivation, such as the autotelic principle 
\citep{Steels-04}, learning progress \citep{KaplanOudeyer-04}, empowerment 
\citep{KlyubinPolaniEtAl-05}, predictive information 
\citep{DerGuettlerEtAl-08,AyBertschingerEtAl-08}, the free energy principle 
\citep{Friston-10}, homeokinesis \citep{DerMartius-12} and others. These 
models have a range of commonalities: they are free of semantics, 
task-independent, universal and can be computed from an agent's subjective 
perspective. Most of the work related to \glspl{IM} focuses on how they 
create \emph{reasonable} behavior (in some suitable sense) for simulated 
agents. There has been some work in the domain of computer games that 
focuses more explicitly on the relationship between intrinsically motivated 
agents and humans, and how an intrinsic motivation could generate more 
believable Non-Player Characters (NPCs) \citep{MerrickMaher-09}, or produce 
generic companions \citep{GuckelsbergerSalgeEtAl-16} or antagonist behavior 
\citep{GuckelsbergerSalgeEtAl-18}.
So far, \glspl{IM} have been deployed on simulated and physical robots
\citep[e.g.,][]{OudeyerKaplanEtAl-07,DerMartius-12,MartiusJahnEtAl-14},
but, as far as we know, there has been no human-robot interaction study yet 
evaluating the perception of intrinsically motivated robots from the 
perspective of humans.
In this work we use predictive information maximization to implement 
an autonomous, intrinsically motivated robot. We describe the formalism in 
more detail in~\autoref{sec:PI}.

\subsection{Autonomy}
The term \emph{autonomy} is used with multiple meanings~\citep{Boden-08}. 
When we talk about autonomous robots, we merely mean robots that are not 
directly controlled by a human operator, autonomy just being a dimension of 
the experimental design~\citep{huang2004autonomy,4136857}. 
In \gls{SDT}, however, autonomy refers to being in control of one's own 
life, which can be seen as a close enough analogy for living 
systems~\citep{Paolo-04}. 
\gls{SDT} also assumes that there is a drive to maintain this state of 
autonomy, which we do not see in general with autonomous robots. We might 
see autonomy used as the idea that a robot should strive to maintain 
operational autonomy, i.e., not be in need of external help, but it usually 
does not refer to a robot striving to not be controlled by a human. 
Finally, autonomy might also be referring to the concept of self-making or 
self-law-giving, which is closely related to 
autopoesis~\citep{MaturanaVarela-91,FroeseZiemke-09}.
In robots, this is currently only a theoretical 
idea~\citep{SMITHERS199788}, 
but it is often considered necessary for \emph{true} intrinsic motivation. 
Any heteronomy during the development or creation of an agent would 
ultimately make them extrinsic and hence undermine their very nature, i.e., 
computational models of intrinsic motivations on robots are usually put on 
those robots by humans, and are thus actually extrinsic. Computational 
models of intrinsic motivation are an attempt to merely reproduce the 
behavior or functionality of genuinely intrinsic motivation in organism. 
This is also the reason that we talk about \emph{perceived} agency and 
\emph{perceived} autonomy. One idea behind this is that by using those 
models for the robots to \emph{pretend} to be intrinsically motivated, 
humans might indeed perceive the robot as thus. In the following, when we 
talk about intrinsic motivation on the robot we exclusively refer to the 
initial, technical meaning, the computational model that aims to mimic 
intrinsic motivation. The more philosophical underpinnings of autonomy are 
highly relevant to the larger context of this work and indicate that this 
approach is useful even if we develop robots with more extensive autonomy, 
making it a robust approach, even for more self-directed robots in the 
future. Here, the main purpose of this section was to clarify that there 
are different levels of autonomy -- so it is clear that when we talk about 
intrinsically motivated autonomy, we do not just speak of a robot that can 
move by itself, but one that can self-directly change its behavior, based 
on a goal that is at least aligned with its own agency. 

\subsection{Predictive Information}
\label{sec:PI}
This section describes \gls{PI} maximization, the intrinsic motivation 
model used for the robot behavior generation in our experiments.
\Gls{PI} has been described as early as~\citeyear{Grassberger-86}, termed 
\emph{effective measure complexity} \citep{Grassberger-86} or \emph{excess 
entropy}~\citep{CrutchfieldYoung-89}. %
Previous work with \gls{PI}-driven robots in simulation demonstrated its 
applicability to a large range of different robot morphologies 
\citep{DerGuettlerEtAl-08,MartiusDerEtAl-13,ZahediMartiusEtAl-13,MartiusJahnEtAl-14}.
A range of existing videos from experiments in 
simulation showcase apparent exploratory, playful, and open-ended behavior 
of individual robots and robot collectives~\citep[see][]{playful-videos-20}.
The \gls{PI}-induced behavior in the videos suggests \gls{PI} as a 
promising immediate candidate measure to test our core idea.

Conceptually, when this measure is transformed into a behavior-generating 
rule, the resulting dynamics essentially fall into a family of learning 
rules related to the reduction of the time prediction error in the 
perception-action loop of a robot \citep[see especially the book The 
Playful Machine,][]{DerMartius-12}.
The aforementioned book also shows how these approaches can be computed 
from the robot's perspective alone. Additionally, the variety of different 
robots and their behaviors presented there shows how different behaviors 
arise from the same formalism due to the sensitivity towards the agent's 
specific embodiment.

The \glsdesc{PI} formalism consists of computing a specific learning rule 
that aims to maximize the mutual information between a robot's past and 
future sensor states~\citep{AyBertschingerEtAl-08}, i.e., \gls{PI} 
quantifies how much information a history of past sensor states contain 
about future sensor states.
More generally, predictive information is defined as the mutual information 
between the past and the future of a robot's sensor input. A high amount of 
predictive information requires two things: First, past sensor states 
should make future sensor states more predictable. This should lead the 
robot to act so that its actions have predictable consequences. 
Furthermore, the robot also needs to create a high variety of sensor input. 
If the robot would always perceive the same sensor input, then there is 
either insufficient information in the past to predict future sensor 
states, or an insufficiently varied future for which there is not much to 
predict. In both cases, an impoverished sensor input reduces the predictive 
information. Alternatively, if there is strong variation in the sensor 
input but little structure in the sensor data stream, i.e., the past has 
little to do with the future, that also leads to low predictive 
information. Vice versa, a high value for predictive information 
requires a high entropy in future sensor states, i.e., a richly varied 
future (a robot motivated to \emph{excite} its sensors to reach a rich 
variety of different states) which at the same time depends on the 
observable past (i.e., which the robot can predict well based on the past).
The behavioral regime is created by these two counterpoised requirements: 
predictability and variety.
This yields a robot wanting to act so that its future is highly 
predictable, while exploring and experiencing new sensor states.
The \gls{PI} literature argues that this balancing act produces rich 
exploratory behavior that is sensitive to the robot's embodiment and argues 
that predictive information is ``the most natural complexity measure for 
time series''~\citep{BialekNemenmanEtAl-01,MartiusDerEtAl-13}.

A robot which acts depending on the maximization of \gls{PI} only compares 
its sensors channels on an information theoretic level, without the need of 
pre-defining any meaning to the sensors. The quantity, therefore, falls in 
the characterization of \gls{IM} presented by \citet{OudeyerKaplan-08} 
(see~\autoref{sec:IM}) and it is a candidate measure to enable 
intrinsically motivated autonomy in a robot.

\citet{DerGuettlerEtAl-08} and 
\citet{AyBertschingerEtAl-08,AyBernigauEtAl-12}
presented derivation rules for \gls{PI}, which allows for computing the 
model directly for linear systems with stationary dynamics.
The next section discusses an extension of their work by 
\citet{MartiusDerEtAl-13} for the use in nonlinear and nonstationary 
systems -- such as physical robotic systems.

\section{Time-local predictive information}
\label{sec:TiPI-overview}
The \glsdesc{PI} formalism to generate the robot's intrinsically 
motivated behavior in the studies of this article is closely following the 
implementation of \citet{MartiusDerEtAl-13}.
They propose an approximation to compute \gls{PI} for nonlinear systems 
with nonstationary dynamics, which allows for behavior development of a 
self-determined robotic system.
They approximate PI with assuming small, Gaussian noise and only consider a 
time window over the current state of the robot and $\tau$ steps back in 
the past, coining it \gls{TiPI}.
\gls{TiPI} allows for going beyond discrete finite-state actions, which 
still dominates scenarios of information theory-based behavior generation, 
towards continuous actions. This permits using physical robots in 
high-dimensional state-action spaces.
\gls{TiPI} enables robot behavior with self-switching dynamics in a simple 
hysteresis system and spontaneous cooperation of physical coupled 
systems~\citep{MartiusDerEtAl-13}.

\Gls{TiPI} works by updating the two internal neural networks of the robot, 
one that generates behavior from sensor input and the other that predicts 
the future states.
The continuous adaptation, aimed at improving the TiPI, moves the robot 
through a range of behavioral regimes.
Importantly, the changes in behavior are partially triggered by the 
interaction with the environment, as mediated through the robot's 
embodiment.
The rate at which those internal neural networks are updated is the
one model parameter which could be adapted for individual 
preferences~\citep{DerMartius-06}.

The approach allows to change the robot's morphology without having to 
redesign the algorithm, but still remaining sensitive to the embodiment of 
the robot, meaning that the resulting behavior differs, depending on how 
the robot interacts with the world.
The morphology can be changed by changing physical parts or by choosing 
different sensors as inputs for the robot's neural networks. In both ways, 
the robot can be guided towards exploring and playing in different ways. 
For example, by including a sensor for the robot's angular velocity around 
its main axis, the spherical robot would try to spin clockwise and 
anticlockwise with changing velocities.
If we further include an accelerometer providing measurements of the 
forward and backward acceleration, the robot would try to explore the 
relationship between spinning movements and locomotion, yielding a variety 
of additional motion patterns. 
If, furthermore, a human is interacting with the robot, this can increase 
the behavioral diversity, depending on the interaction between the robot 
and the human.

\subsection{Deriving update rules}
\label{sec:TiPI-derivation}
\glsreset{TiPI}
\citet{MartiusDerEtAl-13} present estimates of the \gls{TiPI} for general 
stochastic dynamical systems.
For systems with Gaussian noise and with gradient ascent on the \gls{TiPI} 
landscape, they derive explicit expressions for exploratory dynamics. 
This subsection introduces the derivation of the explicit expression.
The derivations are kept short providing only the basic concepts of the 
quantity and introduce the underlying main approximations and assumptions 
that need to be considered when applying the algorithm to a robot in an 
\gls{HRI} scenario. For a detailed treatment, the reader should refer 
to~\citep{AyBertschingerEtAl-08,MartiusDerEtAl-13}. 

Assume a robot has $n$ sensors and the sensor readings are polled in 
constant 
time steps ($\Delta{t}=1$). 
Combine now the result of all sensor values in a vector $s \in \mathbb{R}^n$.
A series of those sensor readings between points of time $a$ and $b$ (with 
$a<b$) 
can be described as a time-discrete process $\{S_t\}^b_{t=a}$, where both 
boundaries are included. %
Let the past be defined by the points of time 
$a,\dots,t-1$ and the future by $t,\dots,b$. 
\citet{BialekNemenmanEtAl-01}
defines the \gls{PI} for some point in time $t$ 
for the time series $S$ as the mutual information between the past and the 
future.
Intuitively, the mutual information measures the shared information of two 
random 
variables, here $S_{\text{past}}$ and $S_{\text{future}}$, i.e., it measures 
how much knowledge of the past $S_{\text{past}}$ reduces the uncertainty of the 
future 
$S_{\text{future}}$. The predictive information, expressed as mutual 
information, is thus defined as follows

\begin{align}
  I(S_{\text{future}}; S_{\text{past}})
  &= \left\langle \ln \frac{p(s_{\text{future}},s_{\text{past}})}{p(s_{\text{future}})p(s_{\text{past}})} \right\rangle \nonumber\\  
  &= H(S_{\text{future}}) - H(S_{\text{future}}|S_{\text{past}})
  \label{eq:PI}
\end{align}
with the average taken over the joint probability density distribution 
$p(s_{\text{past}}, s_{\text{future}})$.

The first essential simplification proposed by 
\citet{MartiusDerEtAl-13} 
is applying the Markov assumption to 
\autoref{eq:PI}.
If $\{S_t\}^b_{t=a}$ is a Markov process, all past information relevant to the future is stored in the 
very last state of the system, i.e., $S_\text{past} = S_{t-1}$.

The predictive information in this case reduces to:
\begin{align}
  I(S_t; S_{t-1})
  &= \sum_{s_{t-1} \in S_{t-1}}\sum_{s_t \in S_t} p(s_t,s_{t-1}) \ln
  \biggl(\frac{p(s_t,s_{t-1})}{p(s_t)p(s_{t-1})}\biggr) \nonumber\\
  &= H(S_t) - H(S_t|S_{t-1})\;.
  \label{eq:PI_markov}
\end{align}

In general, the Markov assumption only holds true for real-world 
sensor processes in exceptional cases. Nonetheless, as in the wide use of, 
e.g., particle or Kalman filters, it is a popular assumption for successfully 
approximating problems using a Bayesian 
approach~\citep{ThrunBurgardEtAl-05}.
\citet{MartiusDerEtAl-13} 
use the reduced \autoref{eq:PI_markov} as the 
definition of the objective function for deriving the autonomous exploration 
dynamics.

\autoref{eq:PI_markov} is a quantity derived for the whole process. However, to 
create an actual behavior rule that reacts to the current situation, it 
necessary to compute a local quantity, specific to the current situation.
Therefore, instead of computing the probability distribution $p(s_t)$ over 
the whole process, we additionally condition the PI on a state $s_{t-2}$. 
The new quantity derived is then

\begin{align}
  I(S_t; S_{t-1} | s_{t-2})
  \label{eq:TiPI_markov}
\end{align}

Because of above Markovianity, this is effectively a time-local quantity 
for PI and therefore it is called \emph{time-local} predictive 
information~(TiPI).
To calculate the \gls{TiPI}, a model of $S_t$ needs to be learned to 
predict its time series. Let $\psi = \mathbb{R}^n 
\rightarrow \mathbb{R}^n$ be a function predicting the time series at 
$t-2$, $t-1$, and $t$ via %

\begin{align}
\hat{s}_{t-2} &= s_{t-2}\\
\hat{s}_{t-1} &= \psi(s_{t-2}, \theta_{t-2})\\
\hat{s}_t     &= \psi(\psi(s_{t-2}, \theta_{t-2}), \theta_{t-1})
\label{eq:forward_model}
\end{align}

In an example implementation~\citep{tipi-implementation}, $\psi$ is 
realized as a one-layer neural network.
$\theta$ is a set of parameters representing the synaptic weights and 
biases that are updated each time step in order to increase \gls{TiPI}.
The actual dynamics of the process can be described via

\begin{equation}
s_t = \psi(s_{t-1}, \theta_{t-1}) + \xi_t
\label{eq:process-dynamics} %
\end{equation}
$\xi_t$ being the prediction error. 

We denote the deviation of the actual dynamics (\autoref{eq:process-dynamics}) 
from the deterministic prediction (\autoref{eq:forward_model}) as

\begin{align}
\delta s_{t'} &= s_{t'} - \hat{s}_{t'}
\label{eq:TiPI_onError_nonlinear}
\end{align}
for any time $t'$ with $t-2 \leq t' \leq t$. Since $s_{t-2}$ is the initial 
state for \gls{TiPI}, there is no deviation at time $t-2$ and $\delta s_{t-2} = 
0$, while one step after the initial state $\delta s_{t-1} = \xi_{t-1}$.
Intuitively, $\delta s_{t}$ represents the prediction error(s) accumulated 
from the start of the prediction (here at $t-2$) up to time $t$.

For very small prediction errors the dynamics of $\delta s$ 
(\autoref{eq:TiPI_onError_nonlinear})
can be linearized as an approximation:

\begin{equation}
 \delta s_{t'} = L(s_{t'-1})\delta s_{t'-1} + \xi_{t'} + O(||\xi_t||^2)
 \label{eq:TiPI_onError_linear}
\end{equation}
with the Jacobian 
\begin{equation}
L_{ij}(s) = \frac{
  \partial \psi_i(s,\theta)
  }{
  \partial s_j
  } \nonumber
\end{equation}

Assuming that the prediction errors $\xi$ are both small and Gaussian, the 
\gls{TiPI} on the deviation process $\delta S_{t'}$ is the same as on the 
original process $S_t$ \citep[see][sec.~A]{MartiusDerEtAl-13b}. %
It is therefore sufficient to concentrate on the error propagation for the 
computation of the \gls{TiPI}.
This reduces \autoref{eq:PI_markov} in such a way that only the probability 
distribution of the deviation $p(\delta s)$ needs to be known, rather than 
the probability distribution over the full state $p(s)$.

If we further assume that the prediction error $\xi$ is white Gaussian, the 
entropy can be expressed as covariances~\citep{CoverThomas-12}. The 
resulting explicit expression of \gls{TiPI} on $\delta S$ becomes:

\begin{equation}
I(\delta S_t ; \delta S_{t-1} | s_{t-2}) = \frac{1}{2} \ln|\Sigma_t| - 
\frac{1}{2}\ln|D_t|
\label{eq:explicit}
\end{equation}
with $\Sigma = \langle\delta{s}\,\delta{s}^T\rangle$ as the covariance 
matrix of the process $\delta 
S$, and $D=\langle\xi\xi^T\rangle$ as the covariance matrix of the 
prediction error.
Note that the predictive information becomes meaningful only at $t$, as the 
prediction error vanishes at $t-2$ and at $t-1$ the two covariance matrices 
coincide: $\Sigma_{t-1} = D_{t-1}$.
The covariances are exact for Gaussianity. For the general case they are 
approximations only.

We now give the algorithm used to drive a robot's behavior towards 
increasing \gls{TiPI}. \citet{MartiusDerEtAl-13} derive it explicitly for 
the gradient ascending neural network presented in 
\autoref{eq:forward_model}.
They argue that the prediction error $\xi$ is essentially 
noise and does not depend on the parameter of the controller, and that 
therefore the term $\ln|D|$ of \autoref{eq:explicit} can be omitted when 
computing the gradient.
Based on \autoref{eq:explicit}, the resulting gradient step executed at 
each time $t$ is

\begin{equation}
 \Delta \theta_t 
 = \epsilon \frac{\partial I}{\partial \theta}
 = \epsilon \frac{\partial}{\partial \theta} \ln|\Sigma_t|
 \label{eq:gradient_ascent}
\end{equation}
with $\epsilon$ being the update rate and $\theta_{t+1} = \theta_t + \Delta 
\theta_t$.

Applying \autoref{eq:TiPI_onError_linear} to above equations results in explicit gradient step

\begin{equation}
\Delta \theta = \epsilon\, \left\langle \delta u^T_t\, \frac{\partial L(s_{t-1})}{\partial 
\theta} \,
\delta s_{t-1} \right\rangle
\label{eq:gradient_ascent_explicit_with_average}
\end{equation}
where $\delta s$ and the auxiliary $\delta u$ are given as 

\begin{align}
\delta s_{t-1} &= s_{t-1} - \psi(s_{t-2}, \theta_{t-2}) \nonumber\\
\delta s_t &= s_t - \psi(\psi(s_{t-2}, \theta_{t-2}), \theta_{t-1}) \nonumber\\
\delta{u} &= \Sigma^{-1}_t\delta{s}_t \nonumber\\
\Sigma_t &= \langle\delta{s}_t \,\delta{s}_t^T\rangle \nonumber
\end{align}

To render $\Delta \theta$ computable, \autoref{eq:gradient_ascent_explicit_with_average} is further approximated by applying the \emph{self-averaging property} (we explain this in more detail below) of a stochastic gradient

\begin{equation}
\Delta \theta = \epsilon\, \delta u^T_t\, \frac{\partial L(s_{t-1})}{\partial 
\theta} \,
\delta s_{t-1}
\label{eq:gradient_ascent_explicit}
\end{equation}

As per \citep{DerGuettlerEtAl-08,MartiusDerEtAl-13},
\autoref{eq:gradient_ascent_explicit} is the equation by which the 
(approximate) \gls{TiPI} 
maximization is ultimately implemented.
We remark that increasing $|\Sigma|$ corresponds to an increase of the norm of 
$\delta{s}$.
In other words, this reflects the amplification of small fluctuations in the 
motor dynamics, i.e., an increase of the instability of the system dynamics.

\subsection{Approximations and assumptions}
\label{sec:TiPI-considerations}
Along with the above derivation, several approximations and assumptions 
have been made. When the measure is applied to a real robot in a real-world 
human-interaction scenario, this requires a careful consideration of the 
assumptions and approximations, which we do in the following.

\subsubsection{Markov assumption}
This assumption simplifies the definition of the objective function 
\autoref{eq:PI_markov}. More importantly, it renders TiPI 
(\autoref{eq:TiPI_markov}) computable as it simplifies the conditional 
probability density distribution.
Applying the assumption to robotics-related problems, especially to make 
Bayesian problems manageable, is common in 
robotics~\citep{ThrunBurgardEtAl-05}. 
This approximation therefore can be considered a popular robotics strategy 
for applying information theory and Bayesian algorithms to the real world.

\subsubsection{Conditioning on an initial state from two states back}
To compute PI for nonlinear systems with nonstationary dynamics, the 
proposed solution is to condition the quantity on an initial state from two 
steps back in time. 
We stick here to the minimal possible window mainly because computing a 
larger window online comes to a computational cost challenging to bear on 
embedded systems.

The sensors used for the input need to be meaningful for the time window. 
For example, a global position of the robot does not change much within the 
time window of two steps, so the robot cannot excite the sensor value in 
the chosen window.
It is therefore preferable to choose sensors which display variation within 
the given time window, such as proprioceptive sensors measuring the 
acceleration or velocity.

\subsubsection{Prediction errors are both: very small and Gaussian}
These assumptions are made at various places for deriving the explicit 
update rules. For example, the assumptions were used to show that TiPI on 
the process $\delta{S}$ (propagation of errors) is equivalent to the one on 
the original process $S$ (sensor states).
This enables the linearization of the error dynamics 
\autoref{eq:TiPI_onError_linear} and eventually, under the same 
assumptions, the formulation of explicit TiPI expressions 
(\autoref{eq:explicit}).
Assuming that the error is very small and Gaussian has implications on 
choosing the right sensors for the experiments. 
Therefore, care needs to be taken that the noise of the sensors remains 
somewhat Gaussian and somewhat small for the duration of the time window.
For example, the motor position typically changes in a continuous fashion 
and therefore the respective sensors are good candidates to fulfill this 
assumptions.

On the contrary, it would violate the Gaussianity assumption to use a 
sensor whose values exhibit, e.g., sudden drops, such as proximity sensors 
based on Bluetooth~\citep{ScheunemannDautenhahn-17}. Such sensors measure 
the signal strength to an external device which is prone to occlusions and 
can sometimes intermittently fail to provide any reading at all.
To mitigate this, it is possible to use filters to smoothen the sensor readings.

\subsubsection{Applying the self-averaging property for stochastic gradients}
\autoref{eq:gradient_ascent_explicit} uses the so-called self-averaging 
property of stochastic gradients,
which means that a stochastic gradient over a larger number of steps in a 
sequence acts as an approximation of averaging over the probability 
distribution~\citep{VanRensburgRechnitzerEtAl-01}.
In other words, we can replace the average over multiple independently 
drawn samples by a one-shot gradient.

Practically, this makes \autoref{eq:explicit} computable, as the density 
distribution of the gradient is hard to obtain.
\citet{MartiusDerEtAl-13} note that using this property is 
only exactly valid for a small update rate $\epsilon$ when it is driven to 
zero eventually.
Note that the update rate $\epsilon$ in our application is quite large to 
allow for a very fast adaptation process.
\citet{MartiusDerEtAl-13} argue that the explicit update 
rules favor the approach of getting an ``intrinsic mechanisms for the 
self-determined and self-directed exploration'', with the exploration being 
driven only by the sensor values. Thus, the one-shot nature of the 
gradients favors the explorative nature of the exploration dynamics and 
increases interesting synergy effects, but is not strictly implementing the 
average.

\subsubsection{Noise is independent of the controller parameters}
To derive the explicit update rules (\autoref{eq:explicit}), the covariance 
of the noise $D=\langle\xi\xi^T\rangle$ is omitted altogether.
The propagation in error is only assumed to be pure independent noise in the
environment.
In other words, the noise is independent of the controller 
parameter~$\theta$.
\citet{MartiusDerEtAl-13} justify this because of the 
``parsimonious control'' implemented by the formalism.

All these assumptions are of course no longer strictly valid once the robot 
interacts with the environment, especially humans. Nevertheless, the 
intended richness of the robot's behavior is not hampered by that. Instead, 
the formalism gives rise to a varied and manifold repertoire of behaviors, 
as shown by many studies mentioned 
in~\citep{AyBertschingerEtAl-08,DerMartius-12,MartiusDerEtAl-13,MartiusJahnEtAl-14}.

\subsection{Applicability for an HRI study}
\label{sec:TiPI-summary}
\citet{MartiusDerEtAl-13} apply the above maximization of 
\gls{TiPI} to simulated robots. As a result, those robots show complex 
behavior~\citep{MartiusDerEtAl-13}.
One example is a humanoid robot with 17~degrees of freedom (DoF) controlled 
by a single high-dimensional controller implementing the PI optimization 
principle from \autoref{eq:gradient_ascent_explicit}. Importantly, despite 
using the same rules, the formalism produces different behavioral regimes 
of the simulated humanoid, depending on the environment it is exposed to.
Its universality for different embodiments and nonstationary settings makes 
it a good candidate for applying it to a robot without concerning oneself 
too much with the environment or the robot's particular embodiment.
However, this formalism on its own does not result in high-level behavior, 
such as walking or serving a human, which may be directly usable in 
traditional \gls{HRI} scenarios. 
It is not a general control algorithm which may generate extrinsic
motivation, such as helping a human to get its battery charged.
This raises the question of how to align this kind of 
behavior generation with behavior a designer of, e.g, an 
assistance robot would want? 
Both issues can be addressed by combining \gls{IM}-based 
behavior generation with scripted behavior, or behavior based
on extrinsically rewarded reinforcement learning.
In the study presented in the next section, we deliberately
leave these things out and focus on an empirical study of the
effect of intrinsic motivation on its own.
In other words, we intend to investigate the effect of
intrinsically motivated autonomy in isolation from additional
criteria and methods for behavior generation.
This evaluation is completely missing from the existing body of work on 
\gls{TiPI} or intrinsically motivated autonomy in general.
This is the gap this article aims to fill.

\section{Materials and methods}
\label{sec:study_design}
This section describes the materials and methods. Note that design choices 
have been motivated by lessons learned from our preliminary 
study~\citep[Chapter~2]{ScheunemannSalgeEtAl-19,Scheunemann-21}. We point 
out the differences to the present study where relevant.

\subsection{Robot}
\label{sec:robot}
\begin{figure}[htb]%
  \centering%
  \begin{subfigure}{.3\linewidth}%
    \centering%
    \subcaption{The robot platform}%
    \includegraphics[width=\linewidth]{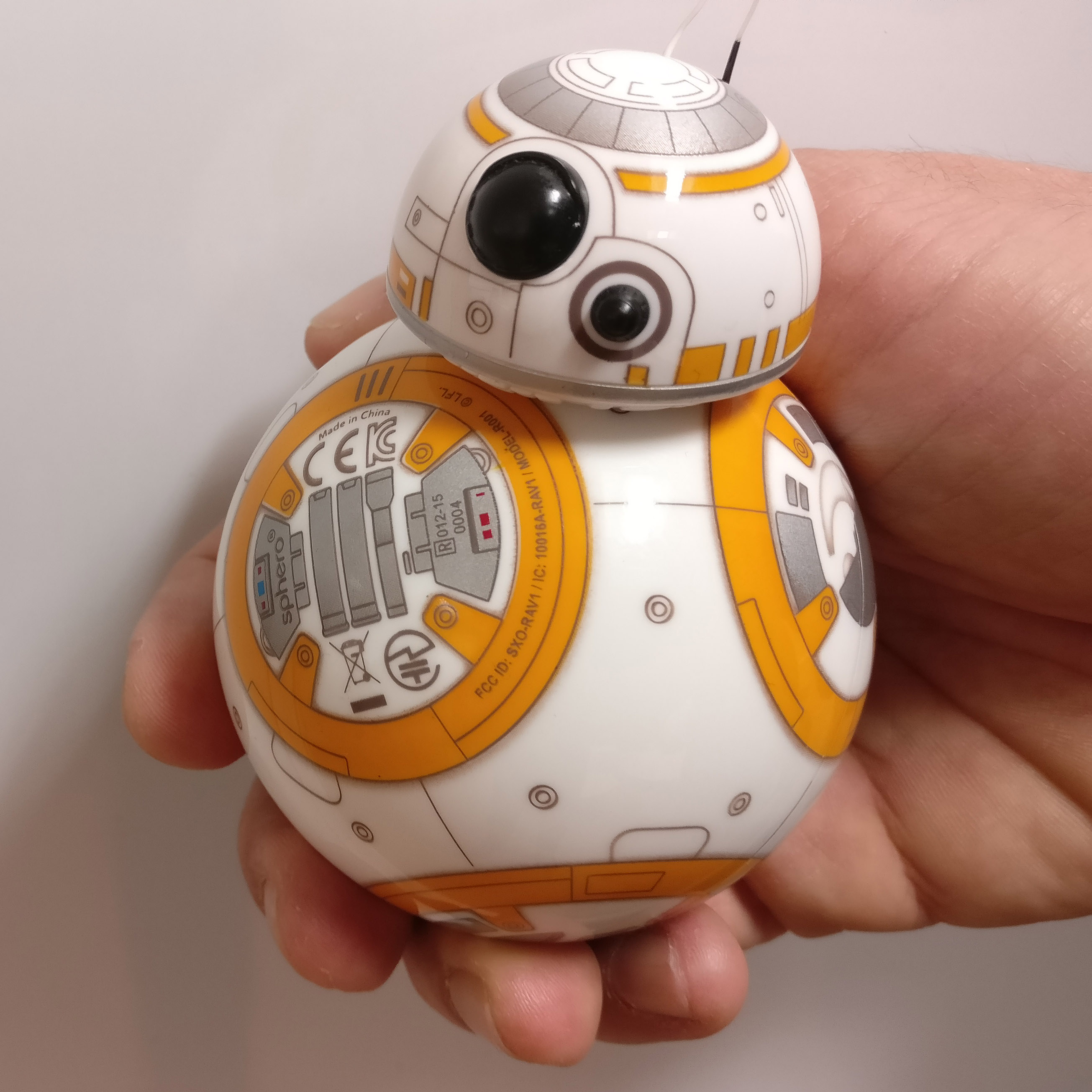}%
    \label{fig:robot:img}
  \end{subfigure}%
  \begin{subfigure}{.3\linewidth}%
    \centering%
    \subcaption{Schema}%
    \includegraphics{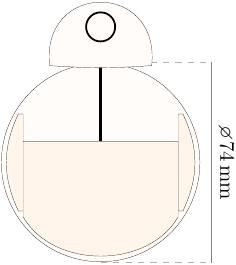}%
    \label{fig:robot:schema}%
  \end{subfigure}%
  \caption{\subref{fig:robot:img}: The used robot platform BB8 from Sphero.
    \subref{fig:robot:schema}: A 2-D cross-sectional view of the robot. A 
    two-wheel vehicle (darker shape), kept in position by a heavy weight, 
    moves the sphere when driving. The speed of each servo motor can be set 
    individually, allowing the robot to move straight, to turn and to spin. 
    A magnet attached to the vehicle keeps the head on top of the sphere 
    facing towards the movement direction.}
  \label{fig:sphero}
\end{figure}

We use the off-the-shelf spherical robot from the company 
Sphero, specifically, the BB8 platform~\citep{sphero-bb8},
as depicted in \autoref{fig:sphero}.
BB8 is a character from the \emph{Star Wars} 
movies~\citep{LucasfilmTwentiethCenturyFoxHomeEntertainment-15}.
A magnet keeps the head in the driving direction, which gives the user a 
sense of the robot's direction. We believe that this helps the human 
participants to interact with the robot.

A humanoid robot platform may raise expectations of advanced social 
capabilities in participants~\citep{Dautenhahn-04,HayashiShiomiEtAl-10}. 
For example, humans would expect the robot to have similar eyesight as 
themselves, or be able to speak and gesture similarly. This in turn may 
interfere with the investigation of how \gls{IM} is being perceived. We 
wanted to reduce as many factors as possible to maximize the focus on the 
effects induced by \gls{TiPI} solely, which is why we decided to use a 
non-humanoid platform with few degrees of freedom.

The robot's on-board hardware is proprietary which prevented us from 
flashing it or running our code. However, it is possible to communicate 
with the robot using \gls{BLE}. 
We can, for example, request a stream of sensor information from the robot, 
or control the robot by either using (i) the robot's balance controller or 
(ii) directly setting the speed of each servo.

The balancing controller receives speed and heading as input values. The 
heading is globally initialized to zero degrees when the robot is started. 
This means, on sending \SI{20}{\degree} to the controller, the robot always 
sets its heading towards 20 degrees on top (clockwise turn) of the initial 
heading.
This controller is a closed-loop controller. If the robot gets nudged or 
turned, it tries to keep the previously set heading constant. This is a 
closed-loop controller since the resulting servo speed depends on the 
readings of the robot's \gls{IMU}.

It is also possible to directly set the speed of the left and right servo. 
This happens in an open-loop fashion, always setting the speed without any 
further observations.

As for sensors, the robot offers raw sensor information from a 3-axis 
accelerometer, a 3-axis gyrometer and the actual motor speed of each servo 
measured as voltage of the \emph{back electromotive force} (back EMF).
The robot can stream data from an \gls{IMU} represented in quaternions or 
Euler angles.
Additionally, it offers velocity information along a plane in the x and y 
direction, and also positional data (i.e., odometry) estimated from its 
starting position.
For our studies, we use sensor data from the \gls{IMU}, the accelerometer, 
the gyrometer and the speed of the wheels~(see 
\autoref{sec:studyII:condition}).

We built a custom API to control the robot. The API is based on 
\texttt{C++} and can run on embedded hardware. It communicates with the 
robot using \gls{BLE} to send commands and to read sensor 
data~\citep[see][]{SPHEROCPPrep}.

\subsection{Measures}%
\label{sec:studyII:measures}%
To investigate whether human participants perceive an intrinsically 
motivated robot as a social other, we employed measures from social 
cognition.
There is a long history in the field of social cognition to understand 
human impression formation of other peers based on two central 
dimensions~\citep[e.g.,][]{RosenbergNelsonEtAl-68,WojciszkeBazinskaEtAl-98}.
\citet{WojciszkeBazinskaEtAl-98}
found that two central dimensions explain on 
average \SI{82.25}{\percent} of the variance of impression formation.
A popular model for human impression formation is the stereotype content 
model~(SCM) introduced by \citet{FiskeCuddyEtAl-07}.
According to \citet{FiskeCuddyEtAl-07}, people perceived as 
warm and competent elicit uniformly positive emotions, are in general more 
favored, and experience more positive interaction by their peers. The 
opposite is true for people scoring low on these dimensions, meaning they 
experience more negative interactions.
Warmth and Competence, together, almost entirely account for how people 
perceive and characterize others~\citep{FiskeCuddyEtAl-07}. 

Highly simplified, perceived Warmth leads to positive social bias,
referred to as active facilitation~\citep{CuddyFiskeEtAl-07}.
The Competence dimension mostly moderates this effect. High Warmth and high 
Competence result in admiration, while high Warmth and low Competence 
result in pity~\citep{JuddJames-HawkinsEtAl-05}. The corresponding effects 
for low Warmth are envy and contempt. As a result, Warmth can be considered 
the primary factor for predicting the valence of interpersonal 
judgments~\citep{FiskeCuddyEtAl-07,AbeleHaukeEtAl-16}. This means, it 
primarily predicts whether an impression is positive or negative.

Recent research in \gls{HRI} has shown that human participants prefer to 
interact again with the robot when they perceive its behavior as 
\emph{more} warm compared to another robot 
behavior~\citep{OliveiraArriagaEtAl-19,ScheunemannCuijpersEtAl-20}.
Our expectation, which would need to be confirmed in future work, is that 
the behavior of a robot that is perceived as more warm is a good candidate 
to facilitate long-term interaction.

In our present study we use the \gls{RoSAS} designed by 
\citet{CarpinellaWymanEtAl-17}, 
which tests for the two aforementioned central dimensions \emph{Warmth} and 
\emph{Competence}. %
We accompany the scale with the Godpseed scale~\citep{BartneckKulicEtAl-09}
because (i) it is a popular scale which may allow to compare the results
to other studies and (ii) it encompasses the dimension Perceived 
Intelligence, which is highly related to the Competence 
dimension~\citep{CarpinellaWymanEtAl-17}.
The latter is important because we designed our study in such a way that 
participants should perceive one robot as similarly competent as the other 
robot since the perception of Competence can influence the perception of 
Warmth~\citep{FiskeCuddyEtAl-07}. However, finding evidence for that, i.e., 
for a small or no effect on the Competence
dimension, can be prone to errors. %
We therefore investigate whether two related dimensions, Perceived 
Intelligence and Competence, yield similarly small effects. In case they do 
support each other, it is unlikely that we missed an effect.

The Godspeed scale uses a 5-point semantic differential scale and 
investigates the dimensions \emph{Anthropomorphism}, \emph{Animacy}, 
\emph{Likeability}, \emph{Perceived Intelligence} and \emph{Perceived 
Safety}. The \gls{RoSAS} tests for the dimensions \emph{Warmth}, 
\emph{Competence} and \emph{Discomfort}. \citet{CarpinellaWymanEtAl-17}
do not recommend a specific size for the Likert-questions, but recommend 
including a neutral value, e.g., by having an uneven number of possible 
responses. Our questionnaire consists of 7-point Likert-type items.

This article focuses on the evaluation and discussion of the dimensions 
Warmth, Competence and Perceived Intelligence to answer the research 
questions~(see~\autoref{sec:RQ}).
However, our questionnaires encompass all the dimensions offered by the
\gls{RoSAS} and the Godspeed scale.
The idea is (i) to hide the intent of our questionnaire and (ii) to allow 
future studies to compare their results with our study. We report the 
effects of all dimensions. However, we only discuss our target dimensions 
and any other statistically significant effect which was revealed. This 
allows us to discuss possible implications of unexpected effects 
(warranting the formulation of hypotheses and conducting a new study).

Note that we continue to capitalize the dimensions to indicate that we 
refer to, e.g., the questionnaire dimension Competence, as opposed to true 
competence. At times, however, we use the adjectives, e.g, competent, 
where it is clear that we refer to the dimension.

\subsection{Robotologist study design}
\label{sec:studyII:robotology}
To evaluate the effect of intrinsically motivated behavior generation on human perception of the robot we had to develop a new study design that could fulfill the following criteria:
\begin{enumerate}
    \item\label{itm:two} Encourage the human to pay attention to the robot.
    \item\label{itm:one} Encourage the human participant to interact -- ideally physically -- with the robot.
    \item\label{itm:three} Do not provide an explicit task assignment for the robot to the human.
    \item\label{itm:four} Do not provide a joint or human task assignment that leads to an implicit assumption about the robot’s task.
\end{enumerate}
The aim of the study design is to engage the human participant and to direct 
their attention towards the robot, so that participants can judge the robot 
behavior based on their interaction with the robot. We also want physical 
interaction, so we can test that the effect we are looking at is robust in 
regards to being perturbed by human interaction. This is because it would 
be counterproductive to identify some robot behavior that could encourage a 
human to physically interact with the robot and then have this behavior 
destroyed by the resulting interaction.
In general, there are a lot of task-based study designs that can fulfill 
criteria \ref{itm:two} and \ref{itm:one}, but the challenge was to find 
a design that can fulfill criteria \ref{itm:three} and \ref{itm:four} at 
the same time. 
Criteria \ref{itm:three} and \ref{itm:four} are necessary because our 
research questions focus on how warm the participants' perceive the robot, 
something that can be biased by the perception of the robot's Competence. 
Perceived competence in turn tends to be influenced by how well a robot 
fulfills a stated task -- or even just an implicitly assumed task.

For example, in the preliminary study~\citep{ScheunemannSalgeEtAl-19}, we 
had asked the participants to use their hands to keep the robot from 
falling off the table, to encourage physical interaction. However, this 
likely led to an implicit assumption that the robot should stay on the 
table, and resulted in those robots which were more likely to fall off the 
table being considered less competent. In this study we wanted to minimize 
the explicit and implict task assignment to evaluate the effect of \gls{IM} 
behavior generation on Warmth in the least biased way.

In order to achieve the aforementioned requirements, we designed our study 
so the participants effectively would learn to become \emph{robotologists} 
in analogy to how anthropologists or naturalists would study animals or 
humans. 
The participants' task was to determine if the two presented robot 
behaviors behave the same or different. 
We hypothesized that this encourages interaction with the robot because 
participants usually want to perform well in order to satisfy the perceived 
needs of the researcher~\citep{Orne-62}.
At the same time, however, this task does not influence the 
participants' expectations of the robot directly and it does not create the 
expectation that the robot will help or hinder this assessment. 
We hypothesized that this robotologist study design encourages the 
human to interact with the robot, while at the same time reducing their 
preconceptions.

\subsection{Environment and Interaction}
\label{sec:environment_tasks}
\begin{figure}%
  \centering%
  \includegraphics[width=.6\columnwidth]{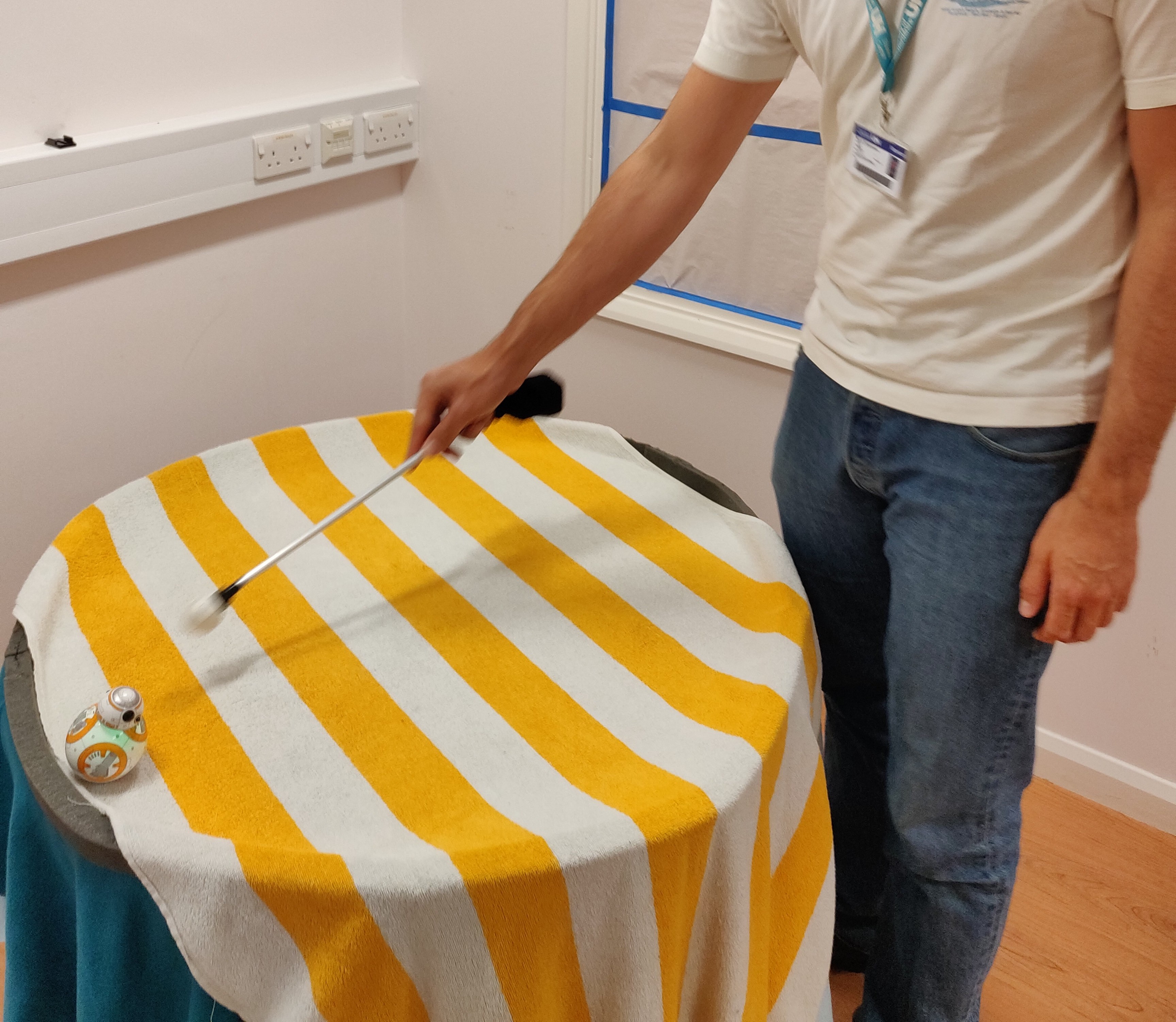}%
  \caption{The picture shows the first author using the interaction tool. 
  He nudged the robot with the white end of the wand. Participants were 
  able to freely chose a position around the table for observing or 
  interacting with the robot.}
  \label{fig:studyII:environment}
\end{figure}
\Autoref{fig:studyII:environment} shows the table that the robot 
locomoted on.
It is circular, with \SI{91}{\centi\metre} in diameter and 
\SI{72}{\centi\metre} in height. 
A foam wall of \SI{2.5}{\centi\metre} in height and with 
\SI{4}{\centi\metre} in width surrounds the border of the table. We decided 
on these measurements in such a way that the robot cannot fall off the 
table, even with a very high velocity. Three blankets of a total height of 
\SIrange{3}{4}{\milli\metre} cover the surface (including the walls). This 
applied some friction and made it easier for the robot to locomote on the 
otherwise smooth and slippery surface of the wooden table top.
The table's distance to the surrounding wall of the room was at least 
\SI{60}{\centi\metre}, which allowed participants to freely move around the 
table.

This study design is different from our preliminary 
study~\citep{ScheunemannSalgeEtAl-19}. There, we positioned the participant 
at one particular side of the table which had no border and therefore 
permitted the robot to fall off the table.
We hypothesized that this encourages interaction with the robot and that 
participants would see that the intrinsically motivated robot adapts to 
their input. Furthermore, we hypothesized that an intrinsically motivated 
robot, which explores the area and the interaction with the participant 
would be perceived as having higher agency and competence.
This design was not entirely successful. The intrinsically motivated robot 
could not sense the edge. Therefore, when it approached participants they 
did not consider it as approaching them, but judged the robot as rather 
\emph{suicidal}. 
In consequence, the participants judged the robot as less competent, 
despite its ability to adapt~\citep[cf.][sec.~5]{ScheunemannSalgeEtAl-19}. 
This is a good example of why it is critical to consider the participants' 
expectations when choosing a robot and designing a study. 

In this study, we decided to keep the robot hardware, but change the design 
of the environment, so that all borders of the interaction environment are 
enclosed. If the participant decides to be passive, the robot cannot fall 
off the table.
In addition, the round shape of the table and its position allows 
participants to reach all borders. We do not assign a specific position to 
the participants. 
This further reduces our instructions to the participants, aiming to 
further reduce our influence on their implicit task assignment for the 
robot.
This design of the interaction environment is key to enable the 
robotologist study design and supports the requirements outlined above in 
\autoref{sec:studyII:robotology}.

\autoref{fig:studyII:environment} shows the first author of this article 
interacting with the robot using the \gls{HRI} tool referred to as a 
\emph{wand} that was developed specifically for this study.
Participants were asked to use the wand to touch and nudge the robot with 
the white end.
The wand is \SI{50}{\centi\meter} long and weighs \SI{78}{\gram}.
It consists of a \SI{40}{\centi\metre} long aluminum tube with a diameter 
of \SI{10}{\milli\metre}. The end is a round, softer sphere. It is made of 
an off-the-shelf table tennis ball with a diameter of \SI{40}{\milli\metre}.

The style of interaction is another difference to the preliminary 
study~\citep{ScheunemannSalgeEtAl-19} where participants used their 
hands to interact with the robot.
The participants in this study were asked to use the wand for interacting. 
We assumed that this would help to ease the interaction, as some 
participants in the preliminary study felt uncomfortable with the idea of 
using their hands for means of interactions. We further hypothesized that 
the mere existence of a tool would make the participants want to use it and 
therefore encourages the interaction.

\subsection{Research questions}
\label{sec:RQ}
The study concerned itself with two main research questions. 
First, we wanted to understand if participants perceive an intrinsically 
motivated robot as a social other and are interested in interacting with 
such a robot. As discussed in \autoref{sec:studyII:measures}, we employed 
the dimension Warmth to measure participants' perception.
We are interested in a \emph{change} of the perception because there is not 
enough research to understand what absolute value of Warmth would be a 
\emph{good} indicator. In this study, we first want to understand whether 
IM-driven behavior has a positive effect:

\begin{description}
  \item[R1:] Is an intrinsically motivated, adaptive robot behavior 
  perceived as more warm compared to a robot with a reactive baseline 
  behavior?
\end{description}%
We hypothesized that we would find evidence that this is the case. We 
did not expect any other strong effects, but we still report and discuss 
the statistically significant main effects of all other dimensions.

In the preliminary study~\citep{ScheunemannSalgeEtAl-19}, we saw evidence 
that participants perceive an intrinsically motivated robot as more warm. 
This, however, needed confirmation because (i) the effect was not 
statistically significant and because (ii) the study design made 
participants perceive one of the robots as more competent than the other. 
This can act as a confounding factor because perceived Competence can 
influence perceived Warmth.
In this study, we aimed to produce a fairer set-up to allow an evaluation 
of Warmth that is not biased by perceived differences of the robot's 
competence.
The above subsections presented what we call the participants' task to 
become a \emph{robotologist} and the environment and the interaction we 
employed in order to achieve that goal.

Drawing a conclusion from \emph{not} finding an effect for Competence is 
prone to the risk that we simply might have missed an effect. To be 
certain, we therefore used the dimension Perceived Intelligence, a 
dimension related to Competence~(see~\autoref{sec:studyII:measures}), and 
expected both of them to show no effect. Our second question is therefore:

\begin{description}
  \item[R2:] Does the robotologist study design help to make the two
  robot behaviors appear to the human participants as similarly competent 
  and intelligent?
\end{description}%
We hypothesized that the answer to the question is ``yes''.
This means that for the dimensions Competence and Perceived Intelligence
we would expect to see no evidence for an effect. 

\subsection{Conditions}%
\label{sec:studyII:condition}%
This experiment consists of two conditions of behavior generation with the 
following characteristics: 
\begin{description}
  \item[\COND{ADA}:] The robot is adapting continuously, based on maximization 
  of \gls{TiPI} and directly applies the resulting \glspl{IM} as servo speed.
  \item[\COND{balanced-REA}:]
  The robot uses its balanced mode for locomotion, the 
  network controlling the robot has been pre-adapted using \gls{PI} and it 
  remains constant.
\end{description}

The reactive robot in the \COND{balanced-REA} condition uses the same 
binary and starts with the same networks as the robots in the preliminary 
study~\citep[cf.][sec.~3.1]{ScheunemannSalgeEtAl-19}. The weights are 
received based on pre-trial adaptation. This determines how it reacts to 
sensor input, but it does not further update its internal network during 
the experiment. 
There are two reasons for taking the \COND{balanced-REA} robot from our 
preliminary study~\citep{ScheunemannSalgeEtAl-19}.
Firstly, the behavior is a good baseline behavior. The robot was 
interesting to the participants and the behavior showed enough variety for 
them to not see any patterns~\citep{ScheunemannSalgeEtAl-19}.
Secondly, keeping the baseline constant, but changing other variables, 
allows for a better comparison to the previous findings and the previous 
intrinsically motivated, adaptive robot.

The intrinsically motivated robot in the \COND{ADA} condition realizes 
behavior motivated by TiPI maximization, and it continuously updates its 
internal networks based on that gradient during the experiment. In contrast 
to the preliminary study, the robot changed the speed of its two servos 
directly, instead of using the balancing 
controller~\citep[cf.][sec.~3.3]{ScheunemannSalgeEtAl-19}. This way the 
robot’s behavior is only influenced by its \glspl{IM}, unconstrained by 
additional software such as the closed-loop balancing controller. In 
particular, this balancing controller might have added meaning to the 
robots output (i.e., staying upright), which may not yield an intrinsically 
motivated robot in accordance to the definition presented in 
\autoref{sec:IM}.
Therefore, this change allowed to further focus the analysis on the 
perception of intrinsically motivated autonomy.

The robot sensor input is again the linear acceleration for the 
forward/backward and left/right axis from the accelerometer, and the 
angular velocity around the upright axis received by the gyrometer.
Instead of using the absolute position of the robot received via its pitch 
and roll angles from the \gls{IMU} as in the preliminary 
study~\citep{ScheunemannSalgeEtAl-19}, we now input the speed of the two 
servos. This allows us to directly couple the output of the controller 
changing the servo speed and the actual measured servo speed.

We wanted both robots to behave similarly at the beginning of the two 
conditions, to avoid the formation of very different first impressions. For 
example, one robot starting off smooth and slow, and the other 
accelerating very fast and bumping into the wall may form a first 
impression in the participant which influences their overall responses.
Therefore, we tweaked the starting weights of the network of the 
intrinsically motivated robot by hand. As there is direct coupling between 
the servo speed readings and the controller output, i.e., the set speed for 
the servos, the weights were set in such a way that a reading on the left 
servo would amplify the output for the left servo, and vice versa.
This way we could create a slow-pace forward movement for the first few 
seconds, which looks similar to the reactive baseline robot. An example 
video of the two conditions of one session accompanies the 
article~\citep[see][]{supp-video}.
As seen in the video, the resulting two robot behaviors look overall very 
similar and are hard to distinguish visually. We believe this is a strength 
of the experimental set-up, as it allows us to rule out other, incidental 
reasons for the observed change in human perception. This makes it more 
likely that it is the adaptation to stimuli received by the robot, driven 
by TiPI maximization, that is responsible for the differences in human 
perception.

\subsection{Procedure}
\label{sec:studyII:procedure}
Participants are welcomed to the experimental room and were then handed an 
information sheet. They were encouraged to discuss concerns related to 
their participation. If they were happy to proceed with the study, they 
were asked to sign an informed consent form.
Then the environment and the robot are presented and briefly described. 
It was then emphasized that they could leave the study whenever they feel 
uncomfortable, stressed or bored.
Participants then complete a pre-questionnaire. This gathers information 
regarding their gender, age and background. 

We then formulated the task for the participants, namely that they should 
find out whether the two presented robots are any different. 
For understanding differences in the robots' behaviors, they can use the 
\gls{HRI} tool: the wand. They are allowed to nudge the robot or block it. 
Both of these actions are demonstrated to the participants. However, no 
other information is provided.

Next, the two conditions are presented to the participants in a randomized 
but counterbalanced order, each lasting approximately \SI{5}{\minute}. They 
complete a post-questionnaire containing the two scales after each condition.
The entire experiment takes \SIrange{50}{60}{\minute} per participant.

\subsection{Participants}
\label{sec:sample}
We recruited 24 participants (10 female; 14 male) mostly from university 
staff and students, between the ages of 18 and 64 years ($M = 31.7,\ 
SD=12.6$). The participants were undergraduates or post-graduates from the 
university, but all na\"ive towards the objectives of the experiment. Eight 
participants had a background in HRI, whereas 9 participants never 
participated in any prior HRI study. 
All were asked how familiar they are with interacting with robots, 
programming robots and the chosen robot platform. 5-point Likert-questions 
were chosen with the value 1 for ``not familiar'' to 5 for ``very 
familiar''.
The self-assessed experience for interacting with robots showed an average 
of 3.5~($\text{Mode}=5$). The average familiarity with programming robots 
3.2~($\text{Mode}=5$) and the experience with the chosen robot platform was 
rated an average 2.1~($\text{Mode}=1$). The familiarity with the movie 
series Star Wars was rated an average of 3.2~($\text{Mode}=4$).

These data were collected primarily to understand what type of participants 
were attracted to the study. We did not form any hypotheses, but we 
expected that many participants may be aware of the Star Wars movies or the 
characters and their impression of the robot's behavior may vary depending 
on their expectations.
However, in this study we are interested whether participants perceive one 
robot behavior as more warm than the other to understand their perception 
of intrinsically motivated autonomy~(cf.~\autoref{sec:RQ}).
This is why we decided for a within-subject design and presented 
both conditions to each participant. This way, we can investigate the 
participants' change in their responses, rather than comparing responses of 
participants with different expectations of the robot's behavior and 
capabilities. This allows us to concentrate on the changes of their responses 
without considering their expectations of the robot's behavior.

The study was conducted on the premises of the University of Hertfordshire 
and was ethically approved by the Health, Science, Engineering \& 
Technology ECDA with protocol number aCOM/PGR/UH/03018(3). The anonymity 
and confidentiality of participants' data are guaranteed.

\subsection{Data analysis}
\label{sec:data_analysis}%
Our research questions asked whether the responses to a questionnaire 
dimension of a participant differ between conditions. In particular, we 
want to know whether participants perceived one robot behavior as more warm 
than the other~(RQ1).
Since we are interested in the change of participants' perception, we 
designed the study so that the robot behavior generation is a 
within-subjects variable, i.e., all its conditions are presented to each 
participant.
This has the benefit that we can answer the research questions with the use 
of statistical tests for pairwise comparisons of the participants' 
responses.

Since our study consists of exactly two conditions, we decided to employ 
two-sample location test of matched pairs: the Wilcoxon signed-rank test.
In contrast to the paired $t$-Test, it is a non-parametric test which is 
both: (i) robust for small sample sizes and (ii) usable for the assessed 
Likert-scale data.
We decided to use the two-sided version of the test to investigate for 
effects in both directions. This way, effects contrary to our hypotheses 
are revealed too.
We use \texttt{wilcox.test(\COND{ADA}, \COND{balanced-REA}, paired=TRUE)} 
to compute the test, a method that is part of \texttt{R}'s built-in 
\texttt{stats} package.

\subsection{Data preparation}
\label{sec:data_preparation}
In order to use the Wilcoxon signed-rank test we first had to analyze the 
data from participants' responses in two steps: (i) we analyzed the scale 
reliability in order to amend scale items if needed and (ii) we tested 
whether the condition responses are independent of their presented order.

\subsubsection{Scale reliability}
To prepare the data for analysis, we tested the score reliability of the 
scales of both standardized questionnaires using Cronbach's $\alpha$. 
\autoref{tab:consistency} presents all test results for the used dimensions.
We found that the item \emph{quiescent--surprised} of the dimension 
Perceived Safety of the Godspeed scale is negatively loaded on the 
dimension. Even if reversed, the reliability is poor with $\alpha=.54$. We 
decided to remove Perceived Safety altogether.
All other dimensions reveal a good score reliability ranging from $.75$ to 
$.85$, or acceptable reliability for the dimension Anthropomorphism: 
$\alpha=.67$.
This is evidence that the scale dimensions are reliable and can be used for 
further investigation.

\begin{table}[htb]
  \centering
  \caption{Internal consistency reliability measured with Cronbach's 
  $\alpha$.}
  \label{tab:consistency}
	\begin{tabular}{@{}llcc@{}}
  \cmidrule[\heavyrulewidth](l){2-4}
  & dimension & items & $\alpha$ \\
  \cmidrule(l){2-4}
  \ldelim\{{3}{4mm}[\parbox{4mm}{\rotatebox[origin=c]{90}{RoSAS}}] 
  & Warmth                 & 6 & .80 \\
  & Competence             & 6 & .85 \\
  & Discomfort             & 6 & .79 \\
  \addlinespace[0.75ex]
  \ldelim\{{5}{4mm}[\parbox{4mm}{\rotatebox[origin=c]{90}{Godspeed}}] 
  & Anthropomorphism       & 5 & .67 \\
  & Animacy                & 6 & .74 \\
  & Likeability             & 5 & .82 \\
  & Perceived Intelligence & 5 & .84 \\
  & Perceived Safety       & 3 & .37 \\
  \cmidrule[\heavyrulewidth](l){2-4}
\end{tabular}
\end{table}

\subsubsection{Interaction effects}
\label{sec:results:interaction}
We then analyzed whether the randomized and counterbalanced assignment of 
the order of conditions to the participants was successful in that the data 
does not show interaction effects between the condition responses and the 
order of the condition. 

\begin{table}[htbp]
  \centering
  \caption{ANOVA-type test results for interaction effects.}%
  \label{tab:studyII:ANOVA}
  \begin{tabular}{llrcc@{}}
  \cmidrule[\heavyrulewidth](l){2-5}
  & dimension &  \multicolumn{1}{c}{$F$} & $df1$ & $p$\\
  \cmidrule(l){2-5}
  \ldelim\{{3}{4mm}[\parbox{4mm}{\rotatebox[origin=c]{90}{RoSAS}}] 
  & Warmth     & 0.001 & 1 & .976 \\
  & Competence & 1.473 & 1 & .225 \\
  & Discomfort & 1.787 & 1 & .181 \\\addlinespace[0.75ex]
  \ldelim\{{4}{4mm}[\parbox{4mm}{\rotatebox[origin=c]{90}{Godspeed}}] 
  & Anthropomorphism  & 0.164 & 1 & .685 \\
  & Animacy           & 0.665 & 1 & .415 \\
  & Likeability       & 1.455 & 1 & .228 \\
  & Perceived Intelligence & \textless0.001 & 1 & .984 \\    
  \cmidrule[\heavyrulewidth](l){2-5}
\end{tabular}
\end{table}
An analysis of variances~(ANOVA) is commonly used for investigating
interaction effects. Following our above argumentation for employing 
non-parametric tests, we use a non-parametric ANOVA-type 
test~\citep{NoguchiGelEtAl-12}.
For computing the ANOVA-type test we used the \texttt{R} package 
\texttt{nparLD}. 
The study can be expressed as F1-LD-F1 Model with the one within-subjects 
variable \emph{behavior generation} (levels: \COND{balanced-REA}, 
\COND{ADA}) and the one between-subjects variable \emph{order} (two 
levels). The \texttt{nparLD} package offers the function 
\texttt{f1.ld.f1()} for computing such models.
\autoref{tab:studyII:ANOVA} shows the results. For a 5\% significance level 
there is no statistical significance and there is not enough evidence for 
an interaction effect for any of the dimensions.

The randomized but counterbalanced ordering of conditions resulted in no 
evidence for interaction effects. Therefore, we can safely investigate the 
main effects independently of their order. This means we can use the 
Wilcoxon signed-rank test and compare the responses to both conditions 
independently of whether the participants were exposed to, e.g., 
\COND{ADA}, at the beginning of the experiment or at the end.

\section{Results}
\label{sec:studyII:results}
Our two research questions asked whether the participants' responses to the 
robot behavior differ between conditions. We therefore employed a paired 
difference test: %
the Wilcoxon signed-rank test. It is a non-parametric test which is known 
to be robust for small sample sizes and can be used for the assessed 
Likert-scale data.
\begin{table*}[htbp]
  \centering
  \caption{Results of the two-sided Wilcoxon signed-rank test for comparing 
  the difference between \COND{balanced-REA} and \COND{ADA}.}%
  \label{tab:studyII:main_effects}
  \begin{tabular}{llrrrrrr@{}}
  \cmidrule[\heavyrulewidth](l){2-8}
  & & & \multicolumn{2}{c}{95\% confidence interval} & \multicolumn{3}{c}{}\\ 
  \cmidrule{4-5}
  & dimension & \multicolumn{1}{c}{$V$} & lower bound & upper bound & 
  \multicolumn{1}{c}{estimate} & \multicolumn{1}{c}{$p$} & 
  \multicolumn{1}{c}{$r$}\\
  \cmidrule(l){2-8}
  \ldelim\{{3}{4mm}[\parbox{4mm}{\rotatebox[origin=c]{90}{RoSAS}}] 
  & Warmth                 &  27.5 & -1.333 & -0.333 & -0.833 & 
  \textsuperscript{*}.007 & .555\\
  & Competence             & 138.5 & -0.833 &  0.583 &  0 & 
  .988                     & .003\\
  & Discomfort             &  54.0 & -1.250 &  0.417 & -0.250 & 
  .287                     & .217\\\addlinespace[0.75ex]
  \ldelim\{{4}{4mm}[\parbox{4mm}{\rotatebox[origin=c]{90}{Godspeed}}] 
  & Anthropomorphism       &  26.0 & -1.200 & -0.500 & -0.900 & 
  \textsuperscript{*}.002  & .635\\
  & Animacy                &  30.5 & -1.250 & -0.417 & -0.833 & 
  \textsuperscript{*}.002 & .636\\
  & Likeability            &  38.0 & -0.700 & -0.100 & -0.400 & 
  \textsuperscript{*}.038 & .424\\
  & Perceived Intelligence & 126.0 & -0.600 &  0.500 & -0.100 & 
  .715                    & .075\\    
  \cmidrule[\heavyrulewidth](l){2-8}
\end{tabular}
\end{table*}

\autoref{tab:studyII:main_effects} shows the results of the two-sided 
Wilcoxon signed-rank test for all dimensions of both scales comparing the 
condition \COND{balanced-REA} and \COND{ADA}.
For each dimension, we report the test statistic $V$, the $p$ value, a 
point estimate and its corresponding confidence interval.
The point estimate (short: estimate) is the median of the difference 
between \COND{balanced-REA} and \COND{ADA}. It provides a magnitude and a 
direction for how much the participants prefer one condition. 
For example, if the estimate for $\COND{balanced-REA}-\COND{ADA}$ equals 
$-0.833$, this means that on average the participants responded to Warmth 
with $0.833$ units higher in the \COND{ADA} condition compared to the 
\COND{balanced-REA} condition. In other words, participants perceive the 
intrinsically motivated robot as more warm than the baseline behavior.
The units here are the responses to the Likert-type scale ranging from 
\numrange{1}{7}~(RoSAS) or the differential scale ranging from 
\numrange{1}{5}~(Godspeed).
Along with the estimate, we further report the standardized effect size $r$.
It allows investigating the size of a potential effect independently of the 
sample size. We follow \citet{Yatani-16} who suggested to use 
Cohen's interpretation: small effect for $r=.1$, a medium effect for $r=.3$ 
and a large effect for $r=.5$.

The results show that there are statistically significant effects for the 
dimensions Warmth~($p=.007$), Anthropomorphism~($p=.002$), 
Animacy~($p=.002$), and Likeability~($p=.038$). The magnitude of the 
effects (estimates) for these effects are negative, which means that 
participants respond higher on each dimension if the robot is intrinsically 
motivated.
We only hypothesized one effect, namely that most participants perceive the 
intrinsically motivated robot~(\COND{ADA}) as more warm than the robot in 
the reactive baseline condition~(\COND{balanced-REA}). 
This directly answers our first research question~(cf.~\autoref{sec:RQ}, 
RQ1), namely that an intrinsically motivated robot (as the one in the 
\COND{ADA} condition) is perceived as more warm.
We will discuss the other unexpected findings in \autoref{sec:discussion}.

\begin{figure*}[htb]%
  \centering%
  \includegraphics{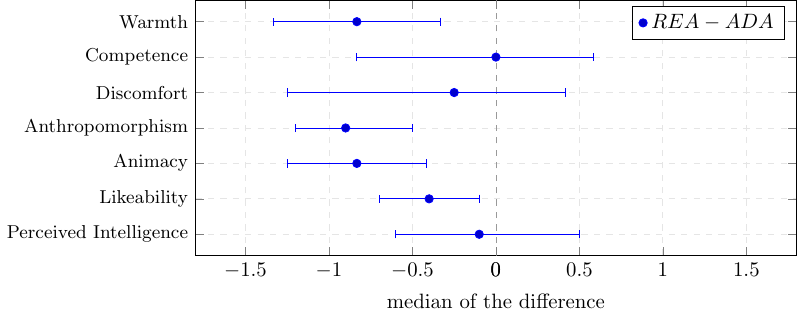}%
  \caption{The figure visualizes the results of the two-sided Wilcoxon 
  signed-rank test reported in \autoref{tab:studyII:main_effects}. The 
  median of the difference (magnitude of the effect) is plotted as point 
  estimates and the 95\% confidence interval as error bars.
  The figure shows that the intrinsically motivated robot (\COND{ADA}) is 
  perceived more warm compared to the baseline robot (\COND{balanced-REA}).
  The figure confirms that there is no magnitude of an effect for either of 
  the two dimensions Perceived Intelligence and Competence.
}
\label{fig:studyII:main_effects}
\end{figure*}

\autoref{fig:studyII:main_effects} visualizes the magnitude of the effect. 
The magnitude of the effect increases with an increasing distance of the 
estimate to zero.
The figure also visualizes the certainty that the point estimate is indeed 
the true effect. The smaller the error bars, i.e., the confidence interval, 
the more certain we can be about the point estimate.
\autoref{fig:studyII:main_effects} confirms that there is a large effect 
for Warmth in favor of the \COND{ADA} condition.

An even larger effect can be observed for the two dimensions 
Anthropomorphism and Animacy. For both dimensions, their perception differs 
and is statistically significant. The estimate and $r>.5$ again indicates 
that there is a large effect in favor of the \COND{ADA} condition. 

The $p$ value for the dimension Likeability is statistically significant 
($p=.038$) and the effect size is $r=.424$ (medium). The magnitude of the 
effect is the smallest among all statistically significant effects. 
Interestingly, it shows the narrowest confidence interval, indicating that 
the magnitude of the effect, while small, is most certain.

As hypothesized, there is no evidence that the robot is perceived 
differently for either of the two dimensions Perceived Intelligence and 
Competence.
From \autoref{tab:studyII:main_effects} we see that the standardized effect 
size $r$ for Competence ($r<.003$) and Perceived Intelligence ($r<.075$) 
indicate that there is no effect.
Moreover, \autoref{fig:studyII:main_effects} shows that the magnitude of an 
effect is almost zero for both of the dimensions. The confidence interval 
is almost equally distributed around zero, indicating that there is no 
certainty for an effect in any direction.
Both dimensions Competence and Perceived Intelligence, which measure a 
similar concept, show a very similar result for testing the perceived 
difference, which provides further support that there is no or only a very 
small effect.
This helps to answer our second research question~(cf.~\autoref{sec:RQ}, 
RQ2), namely whether our robotologist study design helps to make the two
robot behaviors appear similarly competent.

\section{Discussion}
\label{sec:discussion}
The study results provide evidence that the intrinsically motivated robot 
behavior is perceived as more warm than the behavior of the reactive 
baseline robot.
The dimension Warmth is one of the universal dimensions for humans 
judging social attributes of other 
humans~\citep[e.g.,][]{JuddJames-HawkinsEtAl-05,FiskeCuddyEtAl-07,AbeleHaukeEtAl-16}.
Notably, a high scoring for Warmth is considered positive, i.e., desirable, 
and leads to more positive interactions with peers.
There is first evidence that this also applies to robots perceived as more 
warm too. Recent research on robot behavior in HRI has shown that human 
participants prefer to continue interacting with the robot they perceive as 
most warm~\citep{OliveiraArriagaEtAl-19,ScheunemannCuijpersEtAl-20}. 
Our results show that the intrinsically motivated robot displays behavior 
that caused participants to rate the robot higher in Warmth compared to the 
behavior of the reactive baseline robot.
This is an indicator that intrinsically motivated autonomy in robots may 
prove to be relevant for human-robot interaction.
Further support is provided by the statistically significant effect for the 
dimension Likeability, which indicates that participants do like the 
intrinsically motivated robot more.

The current robotologist study design helped us to focus on the Warmth dimension.
Participants did not perceive the robot differently for either of the
two dimensions Perceived Intelligence and Competence in any of the
conditions.
This reflects that the participants did not assign a similar (or any) 
implicit goal to the robot.
Although Competence and Warmth are mainly considered unique
dimensions, some interference and correlations have been pointed out
between them~\citep{FiskeCuddyEtAl-07,AbeleHaukeEtAl-16}. The lack of
an effect for both Perceived Intelligence and Competence is
therefore an important feature of our study design, which allows for
an isolated observation of the influences of IM on the perception of
Warmth. 

We also -- unexpectedly -- observed that participants perceived the 
intrinsically motivated robot as more animated (and as more anthropomorphic).
There is evidence that humans perceive a robot higher in Animacy when the 
robot moves more ``naturally''~\citep{Castro-GonzalezAdmoniEtAl-16}. In 
fact, any object is considered animated if it changes speed and direction 
without visible influences~\citep{TremouletFeldman-00}.
Another influence of the perception of Animacy is the reactivity of the 
robot~\citep{FukudaUeda-10}.
We designed our baseline behavior to provide both: similar movement 
variety, and reaction to sensor input, to allow for a fair comparison and 
focus on the effects of the intrinsic motivation.
In the current study, the control mechanism for the IM robot was 
changed~(cf.~\autoref{sec:studyII:condition}).
Other than the reactive baseline behavior, where the robot could only move 
forward and was kept mostly upright due to the balancing controller, the IM 
robot had a different behavioral regime. It could go backward and forward, 
and because the servo speed was set directly and individually, it could 
produce different behavioral regimes like, e.g., a wobbling locomotion. 
Therefore there are three possible explanations for the baseline behavior 
being perceived as less animated: (i) the motion patterns, (ii) the 
reactivity, or (iii) the intrinsic motivation.
With the current data, we cannot answer this question sufficiently.
Although the baseline behavior was shown to be feasible in the preliminary
study~\citep{ScheunemannSalgeEtAl-19}, we will therefore investigate 
possible changes in a follow-up study, where the baseline will be designed 
to be perceived as similarly animated as the intrinsically motivated robot 
here.

It should be noted that the robot with the baseline behavior is not 
perceived as inanimate in an absolute sense. Instead, participants simply 
perceived the intrinsically motivated robot as more animated and the 
baseline as less animated.
Although this is an indication for the baseline behavior to have less 
natural movements (as discussed), there is no evidence in the literature 
that the rating for Warmth has been significantly influenced.
In the preliminary study \citep{ScheunemannSalgeEtAl-19}, for example, 
participants perceived the baseline behavior as more animated (small 
effect), but they perceived the intrinsically motivated robot as more warm 
(medium effect).
Given the results of both studies, we argue that there is evidence that the
different participant responses for Warmth between the two behavior 
conditions were mainly caused by the robot's intrinsic motivation.

What remains unclear is how much the knowledge from human social cognition 
transfers to human-robot interaction. For example, much like in human 
social cognition, will a robot perceived as warm also experience more 
positive social interaction? Despite recent advances 
\citep[e.g.,][]{OliveiraArriagaEtAl-19,ScheunemannCuijpersEtAl-20}, future 
work has to understand whether the concepts 
from human social cognition transfer to \emph{physical} interaction with a 
robot. If that is the case, our study shows that a robot which has 
intrinsic motivation can help to increase the interest of a human to 
interact with it, 
and that an intrinsically motivated robot is likely to receive more 
positive social interactions.

Our robotologist study design can play a vital role in answering the above 
question, and it can support future research to deepen the understanding 
of the effect of intrinsically motivated robot behavior on the human 
interaction partner.
The study design can be easily amended by (i) using different baseline 
behaviors, (ii) investigating more complex robot behaviors enabled by IM, 
or (iii) using alternative engagement measures.
We already discussed how we plan to compare the intrinsically motivated 
robot with a baseline behavior that is more animated. Other baseline 
behaviors could also include behaviors that are similarly adaptive, but not 
intrinsically motivated.
Once the isolated effect of IM on human perception and its causal 
mechanisms have become more clear, the design can be amended using more 
complex interactions. This could include robot behavior that is not only 
exploratory but also reuses learned behavior. We did not investigate such 
behavior in this study. On the one hand, it is currently an open question 
of how to align IM and learned behavior (more on that later), on the other 
hand, this study serves as first evidence of the specific effect of IM on 
human participants.
Depending on future advances for (i) and (ii), the study design also allows 
using alternative engagement measures for further investigation. 
The current study was chosen to use questionnaires using established 
measures from social cognition as a first step to understand the effect of 
IM. 
However, with increasing behavioral complexity the study would likely need 
to be amended with additional measures for the participant's perception. A 
popular measure for engagement is the time a participant chooses to be 
involved in an interaction with another 
agent~\citep[e.g.,][]{BickmoreSchulmanEtAl-10}. A future 
study could employ the time of interaction as an engagement measure and 
compare the length of the human-robot interaction between conditions. With 
growing interaction complexity, this may need further amendments of other 
engagement measures that investigate distinct interaction parts, such as 
communication or gaze~\citep[e.g.,][]{RichPonslerEtAl-10}. Note that we 
purposefully did not investigate more complex scenarios in this study to 
focus on the effect IM adaptions produce in an even minimal system. This 
simplicity in our study design, in addition, was chosen to limit the 
possibility of alternative explanations for the observed effect other than 
the specific effect between the two conditions.

Coming back to the future plans to investigate more complex human-robot 
interaction scenarios.
We already know from recent research that more complex, anthropomorphic 
robot platforms that \emph{mimic} being intrinsically motivated (e.g., 
curious), are more engaging for 
humans~\citep[e.g.,][]{GordonBreazealEtAl-15,CehaChhibberEtAl-19}. To close 
the gap between the robots \emph{mimicking} \gls{IM}, i.e., with behavior 
implemented and designed by humans, and intrinsically motivated robots 
\emph{per se}, several tasks need to be addressed first. Two main concerns 
here are scalability and alignment. We know from related work that TiPI, 
and other intrinsic motivations, can be scaled up to more complex 
behaviors. For example, \citet{MartiusDerEtAl-13}
presented TiPI-controlled simulated humanoids that generate different 
patterns of behavior in different environments.
This might not be seen as a complex scenario in the field of \gls{HRI} with 
its focus on reliable task completion. This raises the question of how to 
align this kind of behavior generation with behavior a designer would want, 
such as a robot assisting a human? Both issues can be addressed by 
combining \gls{IM}-based behavior generation with scripted behavior, or 
behavior based on extrinsically rewarded reinforcement 
learning~\citep[see][for broadly related initial 
research]{SinghBartoEtAl-04,JaquesLazaridouEtAl-19}.
This can provide a complex scaffolding of behaviors, which then gets 
further enhanced towards more perceived Warmth, by using \glspl{IM}, such 
as TiPI, for behavior modification and local behavior generation. In this 
article, we deliberately left these things out, to focus on an empirical 
study of the effect of intrinsic motivation on its own. In other words, we 
investigate the effect of \gls{IM} in isolation from other factors.

One challenge that this approach poses from the perspective of traditional 
HRI research, is the question of how to identify the isolated factors, such 
as salient behavioral patterns, that cause the robot’s behavior to be 
perceived as more warm.
One motivation for this could be, for instance, that the two resulting 
robot behaviors are hard to distinguish visually. This contrasts with more 
established approaches, where a specific behavior is the tested condition, 
thus providing a more proximal explanation of how certain human perceptions 
were elicited~(see~\autoref{sec:studyII:condition}).
We note that there is research in HRI that does exactly that: it scripts 
robots to display behavior associated with various intrinsic motivations, 
and measures their effect on 
humans~\citep[e.g.,][]{GordonBreazealEtAl-15,CehaChhibberEtAl-19}.
By design, such scripts constitute isolated factors.
Our study does not do this, instead, it investigates the perception of a 
robot where an intrinsic motivation model actually generates and adapts the 
behavior. 
The robot is not just going through the motions of behaving \emph{as if} it 
is curious, it actually seeks out new and predictable stimuli for its 
sensors.
Nevertheless, we do still establish two clearly distinct conditions, 
defined by our controller and the mathematical properties they operate 
under, on which we performed an interventional, double-blind test with 
significantly different results. It would be, of course, of great interest 
to us to better understand the resulting, salient behavior differences that 
provide a proximal, mechanical explanation for these different perceptions. 
This would help to close the gap between IM generation and more scripted or 
task-focused approaches. 
However, we emphasize that using classic HRI techniques for this study to 
understand isolated behavior patterns is a task far from trivial.
This is because of the core property of the information-theoretic measure 
for IM: universality.
Robots that are truly intrinsically motivated can cope with a broad range 
of changes to an agent's environment or its morphology 
(see~\autoref{sec:IM}, 
\ref{sec:PI}, and \ref{sec:TiPI-summary}). We argued that this is key to 
enable robots to interact with humans despite complex environments and 
unknown human characteristics. 
However, it is this precise property that makes it inherently hard to 
investigate for isolated behavioral factors.
Firstly, our exclusive focus on the robot's intrinsic motivation comes with 
the limitation that we cannot, inherently, impose specific behaviors or 
motivations as isolated factors. Doing so would make the robot's behavior 
externally motivated, which is something we did not set out to investigate 
in the present study.
Secondly, information decomposition of the TiPI maximization that is used 
to generate IM behavior could ultimately help to identify and isolate 
factors of the behavior generation. However, the information-theoretic work 
on the question of information decomposition, which has been under intense 
investigation in the last decade, makes it clear that the question on how 
to systematically disentangle different contributions to a truly complex 
behavior is far from straightforward to answer at this 
stage~\citep[e.g.,][]{AyPolaniEtAl-19,RosasMedianoEtAl-20}.
As we can neither control the robot behavior externally nor decompose the 
information that yields the robot's behavior, we would need to research 
methods that extract isolated factors.
We know that observing robot behavior and being part of an interaction with 
the same robot behavior can result in different 
perceptions~\citep[e.g.,][]{FukudaUeda-10}.
Therefore, the employed methods should be more than just extracting 
isolated factors from the robot behavior, but rather be a careful treatment 
of the influence of the robot's behavior on the human and vice versa. This 
means a detailed understanding of the interaction loop between both the 
robot and the human is needed.
Suitable measures to address this satisfactorily do not yet exist. In fact, 
this is why most HRI research uses scripts or teleoperators that mimic 
certain aspects (such as IM). These scripts or the actions of a 
teleoperator do already constitute isolated factors.
We argue that our approach is an important addition to this technique as it 
allows the investigation of fully autonomously generated behavior that 
cannot currently be decomposed; something that, to the best of our 
knowledge, has not yet been researched in HRI.

Despite those long-term challenges, our findings offer some direct 
applications to more current robots.
The here presented TiPI formalism can be used to implement a generic motion 
for situations when a robot does not exhibit a specific behavior, i.e., no 
human is interacting with the robot. We provided evidence that this may 
attract more humans due to their perception of the robot behavior being 
friendly (Warmth).
This would reduce the number of times researchers need to hand-tweak 
natural or affective behavior.
To make a robot more engaging and to elicit curiosity in the human 
interaction partners, we discussed that some researchers proposed that 
novel behaviors, or a larger variety, are important.
These designed behaviors (e.g., uttering questions or statements) are often 
randomly chosen in autonomous 
robots~\citep[e.g.,][]{GordonBreazealEtAl-15}. 
We propose that a more naturalistic selection could be applied by using an 
intrinsic motivation measure. Using TiPI directly is not the best 
candidate, as designed behaviors or scripts cannot be represented by a 
continuous variable.
However, TiPI could be used as a reward signal for a selection algorithm 
based on, e.g., reinforcement learning.
Alternatively, researchers could decide for another formalism implementing 
IM, such as empowerment~\citep{KlyubinPolaniEtAl-05}.
We argue that the study design presented here can help to prototype an 
affective behavior, or affective behavior selection, for a variety of IM 
formalisms. 

\section{Conclusion}
\label{sec:conclusion}
\glsresetall
We started this research with the question if intrinsically motivated 
autonomous robots can be beneficial for designing engaging \gls{HRI}.
We conducted a within-subjects study~($N=24$) where participants interacted 
with a fully autonomous Sphero BB8 robot with two conditions covering 
different behavioral regimes: one realizing an adaptive, intrinsically 
motivated behavior and the other being reactive, but not adaptive.
We used \gls{TiPI} maximization as one candidate measure to produce 
IM-based behavior, and produced, to our knowledge, the first study 
quantitatively investigating human perception of intrinsically motivated 
robots.
Of particular interest is the high similarity between both conditions in 
Perceived Intelligence ($r=.075,\ p=.715$) and Competence ($r=.003,\ 
p=.988$), which gives support to our non-task oriented interaction design. 
This was particularly important as Competence ratings can influence the 
perception of Warmth, which is the dimension we focused on in our study.

Our main result is that an intrinsically motivated robot generates behavior 
that was perceived more warm compared to a baseline robot that was not 
intrinsically motivated~($r=.555,\ p=.007$).
The baseline behavior includes both: similar movement and reaction to 
sensor inputs -- meaning that it is highly likely that the difference in 
perception arises from the robot's adaptation to the physical interaction.
The dimension Warmth is, as mentioned previously, an important factor for 
attitude formation in human-human social cognition. However, it is not 
immediately clear if this higher perceived Warmth leads to a positive 
attitude or preferences in human-\emph{robot} interaction.
If future work would demonstrate this, then we believe the formalism 
presented here could be utilized to create a preference or positive 
attitude towards a robot in a large range of \gls{HRI} scenarios.

The open questions going forward are now: Can we confirm the results by 
using a baseline behavior with more similar motion regimes to further 
strengthen the focus on the \gls{IM} of the agent and the interaction?
Does the universal applicability of the formalism also translate into a 
universal, or at least widespread, evocation of Warmth across different 
robot morphologies? Does this effect persist over time? And does a positive 
social attitude lead to more engagement? All these questions are 
empirically testable, and given the positive results here are possible 
directions for future research.

\subsection*{Acknowledgements}
We would like to thank Rebecca Miko for her help with recruiting 
participants and with her help in proof-reading.

\subsection*{Declaration of conflicting interests}
The authors declare that there is no conflict of interest.

\subsection*{Funding}
MS and DP acknowledge support by the socSMCs FET Proactive project [grant 
number H2020-641321]; and KD acknowledges funding from the Canada 150 
Research Chairs Program.


\begin{thebibliography}{}
  
  \bibitem [\protect \citeauthoryear {%
    Abele%
    \ \protect \BOthers {.}}{%
    Abele%
    \ \protect \BOthers {.}}{%
    {\protect \APACyear {2016}}%
  }]{%
    AbeleHaukeEtAl-16}
  \APACinsertmetastar {%
    AbeleHaukeEtAl-16}%
  \begin{APACrefauthors}%
    Abele, A\BPBI E.%
    , Hauke, N.%
    , Peters, K.%
    , Louvet, E.%
    , Szymkow, A.%
    \BCBL {}\ \BBA {} Duan, Y.%
  \end{APACrefauthors}%
  \unskip\
  \newblock
  \APACrefYearMonthDay{2016}{}{}.
  \newblock
  {\BBOQ}\APACrefatitle {{Facets of the Fundamental Content Dimensions: 
  Agency
      with Competence and Assertiveness{\textemdash}Communion with Warmth 
      and
      Morality}} {{Facets of the Fundamental Content Dimensions: Agency with
      Competence and Assertiveness{\textemdash}Communion with Warmth and
      Morality}}.{\BBCQ}
  \newblock
  \APACjournalVolNumPages{Frontiers in Psychology}{7}{}{}.
  \newblock
  \begin{APACrefDOI} \doi{10.3389/fpsyg.2016.01810} \end{APACrefDOI}
  \PrintBackRefs{\CurrentBib}
  
  \bibitem [\protect \citeauthoryear {%
    Ay%
    , Bernigau%
    , Der%
    \BCBL {}\ \BBA {} Prokopenko%
  }{%
    Ay%
    \ \protect \BOthers {.}}{%
    {\protect \APACyear {2012}}%
  }]{%
    AyBernigauEtAl-12}
  \APACinsertmetastar {%
    AyBernigauEtAl-12}%
  \begin{APACrefauthors}%
    Ay, N.%
    , Bernigau, H.%
    , Der, R.%
    \BCBL {}\ \BBA {} Prokopenko, M.%
  \end{APACrefauthors}%
  \unskip\
  \newblock
  \APACrefYearMonthDay{2012}{}{}.
  \newblock
  {\BBOQ}\APACrefatitle {{Information-driven self-organization: the 
  dynamical
      system approach to autonomous robot behavior}} {{Information-driven
      self-organization: the dynamical system approach to autonomous robot
      behavior}}.{\BBCQ}
  \newblock
  \APACjournalVolNumPages{Theory in Biosciences}{131}{3}{161--179}.
  \newblock
  \begin{APACrefDOI} \doi{10.1007/s12064-011-0137-9} \end{APACrefDOI}
  \PrintBackRefs{\CurrentBib}
  
  \bibitem [\protect \citeauthoryear {%
    Ay%
    , Bertschinger%
    , Der%
    , G{\"{u}}ttler%
    \BCBL {}\ \BBA {} Olbrich%
  }{%
    Ay%
    \ \protect \BOthers {.}}{%
    {\protect \APACyear {2008}}%
  }]{%
    AyBertschingerEtAl-08}
  \APACinsertmetastar {%
    AyBertschingerEtAl-08}%
  \begin{APACrefauthors}%
    Ay, N.%
    , Bertschinger, N.%
    , Der, R.%
    , G{\"{u}}ttler, F.%
    \BCBL {}\ \BBA {} Olbrich, E.%
  \end{APACrefauthors}%
  \unskip\
  \newblock
  \APACrefYearMonthDay{2008}{}{01}.
  \newblock
  {\BBOQ}\APACrefatitle {{Predictive information and explorative behavior of
      autonomous robots}} {{Predictive information and explorative behavior 
      of
      autonomous robots}}.{\BBCQ}
  \newblock
  \APACjournalVolNumPages{The European Physical Journal B}{63}{3}{329--339}.
  \newblock
  \begin{APACrefDOI} \doi{10.1140/epjb/e2008-00175-0} \end{APACrefDOI}
  \PrintBackRefs{\CurrentBib}
  
  \bibitem [\protect \citeauthoryear {%
    Ay%
    , Polani%
    \BCBL {}\ \BBA {} Virgo%
  }{%
    Ay%
    \ \protect \BOthers {.}}{%
    {\protect \APACyear {2020}}%
  }]{%
    AyPolaniEtAl-19}
  \APACinsertmetastar {%
    AyPolaniEtAl-19}%
  \begin{APACrefauthors}%
    Ay, N.%
    , Polani, D.%
    \BCBL {}\ \BBA {} Virgo, N.%
  \end{APACrefauthors}%
  \unskip\
  \newblock
  \APACrefYearMonthDay{2020}{}{}.
  \newblock
  {\BBOQ}\APACrefatitle {{Information Decomposition based on Cooperative 
  Game
      Theory}} {{Information Decomposition based on Cooperative Game
      Theory}}.{\BBCQ}
  \newblock
  \APACjournalVolNumPages{Kybernetika}{56}{5}{979--1014}.
  \newblock
  \begin{APACrefDOI} \doi{10.14736/kyb-2020-5-0979} \end{APACrefDOI}
  \PrintBackRefs{\CurrentBib}
  
  \bibitem [\protect \citeauthoryear {%
    Bartneck%
    , Kuli{\'{c}}%
    , Croft%
    \BCBL {}\ \BBA {} Zoghbi%
  }{%
    Bartneck%
    \ \protect \BOthers {.}}{%
    {\protect \APACyear {2009}}%
  }]{%
    BartneckKulicEtAl-09}
  \APACinsertmetastar {%
    BartneckKulicEtAl-09}%
  \begin{APACrefauthors}%
    Bartneck, C.%
    , Kuli{\'{c}}, D.%
    , Croft, E.%
    \BCBL {}\ \BBA {} Zoghbi, S.%
  \end{APACrefauthors}%
  \unskip\
  \newblock
  \APACrefYearMonthDay{2009}{}{}.
  \newblock
  {\BBOQ}\APACrefatitle {{Measurement Instruments for the Anthropomorphism,
      Animacy, Likeability, Perceived Intelligence, and Perceived Safety of
      Robots}} {{Measurement Instruments for the Anthropomorphism, Animacy,
      Likeability, Perceived Intelligence, and Perceived Safety of 
      Robots}}.{\BBCQ}
  \newblock
  \APACjournalVolNumPages{International Journal of Social
    Robotics}{1}{1}{71--81}.
  \newblock
  \begin{APACrefDOI} \doi{10.1007/s12369-008-0001-3} \end{APACrefDOI}
  \PrintBackRefs{\CurrentBib}
  
  \bibitem [\protect \citeauthoryear {%
    Belpaeme%
    , Kennedy%
    , Ramachandran%
    , Scassellati%
    \BCBL {}\ \BBA {} Tanaka%
  }{%
    Belpaeme%
    \ \protect \BOthers {.}}{%
    {\protect \APACyear {2018}}%
  }]{%
    BelpaemeKennedyEtAl-18}
  \APACinsertmetastar {%
    BelpaemeKennedyEtAl-18}%
  \begin{APACrefauthors}%
    Belpaeme, T.%
    , Kennedy, J.%
    , Ramachandran, A.%
    , Scassellati, B.%
    \BCBL {}\ \BBA {} Tanaka, F.%
  \end{APACrefauthors}%
  \unskip\
  \newblock
  \APACrefYearMonthDay{2018}{}{}.
  \newblock
  {\BBOQ}\APACrefatitle {{Social robots for education: A review}} {{Social 
  robots
      for education: A review}}.{\BBCQ}
  \newblock
  \APACjournalVolNumPages{Science Robotics}{3}{21}{}.
  \newblock
  \begin{APACrefDOI} \doi{10.1126/scirobotics.aat5954} \end{APACrefDOI}
  \PrintBackRefs{\CurrentBib}
  
  \bibitem [\protect \citeauthoryear {%
    Bialek%
    , Nemenman%
    \BCBL {}\ \BBA {} Tishby%
  }{%
    Bialek%
    \ \protect \BOthers {.}}{%
    {\protect \APACyear {2001}}%
  }]{%
    BialekNemenmanEtAl-01}
  \APACinsertmetastar {%
    BialekNemenmanEtAl-01}%
  \begin{APACrefauthors}%
    Bialek, W.%
    , Nemenman, I.%
    \BCBL {}\ \BBA {} Tishby, N.%
  \end{APACrefauthors}%
  \unskip\
  \newblock
  \APACrefYearMonthDay{2001}{}{}.
  \newblock
  {\BBOQ}\APACrefatitle {{Predictability, Complexity, and Learning}}
  {{Predictability, Complexity, and Learning}}.{\BBCQ}
  \newblock
  \APACjournalVolNumPages{Neural Computation}{13}{11}{2409--2463}.
  \newblock
  \begin{APACrefDOI} \doi{10.1162/089976601753195969} \end{APACrefDOI}
  \PrintBackRefs{\CurrentBib}
  
  \bibitem [\protect \citeauthoryear {%
    Bickmore%
    , Schulman%
    \BCBL {}\ \BBA {} Yin%
  }{%
    Bickmore%
    \ \protect \BOthers {.}}{%
    {\protect \APACyear {2010}}%
  }]{%
    BickmoreSchulmanEtAl-10}
  \APACinsertmetastar {%
    BickmoreSchulmanEtAl-10}%
  \begin{APACrefauthors}%
    Bickmore, T.%
    , Schulman, D.%
    \BCBL {}\ \BBA {} Yin, L.%
  \end{APACrefauthors}%
  \unskip\
  \newblock
  \APACrefYearMonthDay{2010}{}{}.
  \newblock
  {\BBOQ}\APACrefatitle {{Maintaining Engagement in Long-term Interventions 
  with
      Relational Agents}} {{Maintaining Engagement in Long-term 
      Interventions with
      Relational Agents}}.{\BBCQ}
  \newblock
  \APACjournalVolNumPages{Applied Artificial Intelligence}{24}{6}{648--666}.
  \newblock
  \begin{APACrefDOI} \doi{10.1080/08839514.2010.492259} \end{APACrefDOI}
  \PrintBackRefs{\CurrentBib}
  
  \bibitem [\protect \citeauthoryear {%
    Boden%
  }{%
    Boden%
  }{%
    {\protect \APACyear {2008}}%
  }]{%
    Boden-08}
  \APACinsertmetastar {%
    Boden-08}%
  \begin{APACrefauthors}%
    Boden, M\BPBI A.%
  \end{APACrefauthors}%
  \unskip\
  \newblock
  \APACrefYearMonthDay{2008}{}{}.
  \newblock
  {\BBOQ}\APACrefatitle {{Autonomy: What is it?}} {{Autonomy: What is
      it?}}{\BBCQ}
  \newblock
  \APACjournalVolNumPages{Biosystems}{91}{2}{305--308}.
  \newblock
  \begin{APACrefDOI} \doi{10.1016/j.biosystems.2007.07.003} \end{APACrefDOI}
  \PrintBackRefs{\CurrentBib}
  
  \bibitem [\protect \citeauthoryear {%
    Carpinella%
    , Wyman%
    , Perez%
    \BCBL {}\ \BBA {} Stroessner%
  }{%
    Carpinella%
    \ \protect \BOthers {.}}{%
    {\protect \APACyear {2017}}%
  }]{%
    CarpinellaWymanEtAl-17}
  \APACinsertmetastar {%
    CarpinellaWymanEtAl-17}%
  \begin{APACrefauthors}%
    Carpinella, C\BPBI M.%
    , Wyman, A\BPBI B.%
    , Perez, M\BPBI A.%
    \BCBL {}\ \BBA {} Stroessner, S\BPBI J.%
  \end{APACrefauthors}%
  \unskip\
  \newblock
  \APACrefYearMonthDay{2017}{}{}.
  \newblock
  {\BBOQ}\APACrefatitle {{The Robotic Social Attributes Scale (RoSAS):
      Development and Validation}} {{The Robotic Social Attributes Scale 
      (RoSAS):
      Development and Validation}}.{\BBCQ}
  \newblock
  \BIn{} \APACrefbtitle {{Proceedings of the 2017 ACM/IEEE International
      Conference on Human-Robot Interaction}} {{Proceedings of the 2017 
      ACM/IEEE
      International Conference on Human-Robot Interaction}}\ (\BPGS\ 
      254--262).
  \newblock
  \APACaddressPublisher{New York, NY, USA}{ACM}.
  \newblock
  \begin{APACrefDOI} \doi{10.1145/2909824.3020208} \end{APACrefDOI}
  \PrintBackRefs{\CurrentBib}
  
  \bibitem [\protect \citeauthoryear {%
    Castro-Gonz{\'{a}}lez%
    , Admoni%
    \BCBL {}\ \BBA {} Scassellati%
  }{%
    Castro-Gonz{\'{a}}lez%
    \ \protect \BOthers {.}}{%
    {\protect \APACyear {2016}}%
  }]{%
    Castro-GonzalezAdmoniEtAl-16}
  \APACinsertmetastar {%
    Castro-GonzalezAdmoniEtAl-16}%
  \begin{APACrefauthors}%
    Castro-Gonz{\'{a}}lez, {\'{A}}.%
    , Admoni, H.%
    \BCBL {}\ \BBA {} Scassellati, B.%
  \end{APACrefauthors}%
  \unskip\
  \newblock
  \APACrefYearMonthDay{2016}{}{}.
  \newblock
  {\BBOQ}\APACrefatitle {{Effects of form and motion on judgments of social
      robots' animacy, likability, trustworthiness and unpleasantness}} 
      {{Effects
      of form and motion on judgments of social robots' animacy, likability,
      trustworthiness and unpleasantness}}.{\BBCQ}
  \newblock
  \APACjournalVolNumPages{International Journal of Human-Computer
    Studies}{90}{}{27--38}.
  \newblock
  \begin{APACrefDOI} \doi{10.1016/j.ijhcs.2016.02.004} \end{APACrefDOI}
  \PrintBackRefs{\CurrentBib}
  
  \bibitem [\protect \citeauthoryear {%
    Ceha%
    \ \protect \BOthers {.}}{%
    Ceha%
    \ \protect \BOthers {.}}{%
    {\protect \APACyear {2019}}%
  }]{%
    CehaChhibberEtAl-19}
  \APACinsertmetastar {%
    CehaChhibberEtAl-19}%
  \begin{APACrefauthors}%
    Ceha, J.%
    , Chhibber, N.%
    , Goh, J.%
    , McDonald, C.%
    , Oudeyer, P\BHBI Y.%
    , Kuli{\'{c}}, D.%
    \BCBL {}\ \BBA {} Law, E.%
  \end{APACrefauthors}%
  \unskip\
  \newblock
  \APACrefYearMonthDay{2019}{}{}.
  \newblock
  {\BBOQ}\APACrefatitle {{Expression of Curiosity in Social Robots}} 
  {{Expression
      of Curiosity in Social Robots}}.{\BBCQ}
  \newblock
  \BIn{} \APACrefbtitle {{Proceedings of the 2019 CHI Conference on Human 
  Factors
      in Computing Systems - CHI'19}.} {{Proceedings of the 2019 CHI 
      Conference on
      Human Factors in Computing Systems - CHI'19}.}
  \newblock
  \APACaddressPublisher{}{{ACM} Press}.
  \newblock
  \begin{APACrefDOI} \doi{10.1145/3290605.3300636} \end{APACrefDOI}
  \PrintBackRefs{\CurrentBib}
  
  \bibitem [\protect \citeauthoryear {%
    Christensen%
    \ \protect \BOthers {.}}{%
    Christensen%
    \ \protect \BOthers {.}}{%
    {\protect \APACyear {2016}}%
  }]{%
    ChristensenOkamuraEtAl-16}
  \APACinsertmetastar {%
    ChristensenOkamuraEtAl-16}%
  \begin{APACrefauthors}%
    Christensen, H\BPBI I.%
    , Okamura, A.%
    , Mataric, M.%
    , Kumar, V.%
    , Hager, G.%
    \BCBL {}\ \BBA {} Choset, H.%
  \end{APACrefauthors}%
  \unskip\
  \newblock
  \APACrefYearMonthDay{2016}{}{}.
  \newblock
  \APACrefbtitle {{Next Generation Robotics}} {{Next Generation Robotics}}\
  \APACbVolEdTR{}{\BTR{}}.
  \newblock
  \begin{APACrefURL} \url{https://arxiv.org/abs/1606.09205v1} 
  \end{APACrefURL}
  \PrintBackRefs{\CurrentBib}
  
  \bibitem [\protect \citeauthoryear {%
    Clabaugh%
    \ \BBA {} Matari{\'{c}}%
  }{%
    Clabaugh%
    \ \BBA {} Matari{\'{c}}%
  }{%
    {\protect \APACyear {2019}}%
  }]{%
    ClabaughMataric-19}
  \APACinsertmetastar {%
    ClabaughMataric-19}%
  \begin{APACrefauthors}%
    Clabaugh, C.%
    \BCBT {}\ \BBA {} Matari{\'{c}}, M.%
  \end{APACrefauthors}%
  \unskip\
  \newblock
  \APACrefYearMonthDay{2019}{}{}.
  \newblock
  {\BBOQ}\APACrefatitle {{Escaping Oz: Autonomy in Socially Assistive 
  Robotics}}
  {{Escaping Oz: Autonomy in Socially Assistive Robotics}}.{\BBCQ}
  \newblock
  \APACjournalVolNumPages{Annual Review of Control, Robotics, and Autonomous
    Systems}{2}{1}{33--61}.
  \newblock
  \begin{APACrefDOI} \doi{10.1146/annurev-control-060117-104911} 
  \end{APACrefDOI}
  \PrintBackRefs{\CurrentBib}
  
  \bibitem [\protect \citeauthoryear {%
    Cover%
    \ \BBA {} Thomas%
  }{%
    Cover%
    \ \BBA {} Thomas%
  }{%
    {\protect \APACyear {2012}}%
  }]{%
    CoverThomas-12}
  \APACinsertmetastar {%
    CoverThomas-12}%
  \begin{APACrefauthors}%
    Cover, T\BPBI M.%
    \BCBT {}\ \BBA {} Thomas, J\BPBI A.%
  \end{APACrefauthors}%
  \unskip\
  \newblock
  \APACrefYear{2012}.
  \newblock
  \APACrefbtitle {{Elements of Information Theory}} {{Elements of 
  Information
      Theory}}.
  \newblock
  \APACaddressPublisher{}{John Wiley \& Sons}.
  \PrintBackRefs{\CurrentBib}
  
  \bibitem [\protect \citeauthoryear {%
    Crutchfield%
    \ \BBA {} Young%
  }{%
    Crutchfield%
    \ \BBA {} Young%
  }{%
    {\protect \APACyear {1989}}%
  }]{%
    CrutchfieldYoung-89}
  \APACinsertmetastar {%
    CrutchfieldYoung-89}%
  \begin{APACrefauthors}%
    Crutchfield, J\BPBI P.%
    \BCBT {}\ \BBA {} Young, K.%
  \end{APACrefauthors}%
  \unskip\
  \newblock
  \APACrefYearMonthDay{1989}{}{}.
  \newblock
  {\BBOQ}\APACrefatitle {{Inferring statistical complexity}} {{Inferring
      statistical complexity}}.{\BBCQ}
  \newblock
  \APACjournalVolNumPages{Physical Review Letters}{63}{2}{105--108}.
  \newblock
  \begin{APACrefDOI} \doi{10.1103/physrevlett.63.105} \end{APACrefDOI}
  \PrintBackRefs{\CurrentBib}
  
  \bibitem [\protect \citeauthoryear {%
    Cuddy%
    , Fiske%
    \BCBL {}\ \BBA {} Glick%
  }{%
    Cuddy%
    \ \protect \BOthers {.}}{%
    {\protect \APACyear {2007}}%
  }]{%
    CuddyFiskeEtAl-07}
  \APACinsertmetastar {%
    CuddyFiskeEtAl-07}%
  \begin{APACrefauthors}%
    Cuddy, A\BPBI J\BPBI C.%
    , Fiske, S\BPBI T.%
    \BCBL {}\ \BBA {} Glick, P.%
  \end{APACrefauthors}%
  \unskip\
  \newblock
  \APACrefYearMonthDay{2007}{}{}.
  \newblock
  {\BBOQ}\APACrefatitle {{The BIAS map: Behaviors from intergroup affect and
      stereotypes}} {{The BIAS map: Behaviors from intergroup affect and
      stereotypes}}.{\BBCQ}
  \newblock
  \APACjournalVolNumPages{Journal of Personality and Social
    Psychology}{92}{4}{631--648}.
  \newblock
  \begin{APACrefDOI} \doi{10.1037/0022-3514.92.4.631} \end{APACrefDOI}
  \PrintBackRefs{\CurrentBib}
  
  \bibitem [\protect \citeauthoryear {%
    Dautenhahn%
  }{%
    Dautenhahn%
  }{%
    {\protect \APACyear {2004}}%
  }]{%
    Dautenhahn-04}
  \APACinsertmetastar {%
    Dautenhahn-04}%
  \begin{APACrefauthors}%
    Dautenhahn, K.%
  \end{APACrefauthors}%
  \unskip\
  \newblock
  \APACrefYearMonthDay{2004}{}{}.
  \newblock
  {\BBOQ}\APACrefatitle {{Robots we like to live with?! - a developmental
      perspective on a personalized, life-long robot companion}} {{Robots 
      we like
      to live with?! - a developmental perspective on a personalized, 
      life-long
      robot companion}}.{\BBCQ}
  \newblock
  \BIn{} \APACrefbtitle {{RO-MAN 2004. 13th IEEE International Workshop on 
  Robot
      and Human Interactive Communication (IEEE Catalog No.04TH8759)}.} 
      {{RO-MAN
      2004. 13th IEEE International Workshop on Robot and Human Interactive
      Communication (IEEE Catalog No.04TH8759)}.}
  \newblock
  \APACaddressPublisher{}{{IEEE}}.
  \newblock
  \begin{APACrefDOI} \doi{10.1109/roman.2004.1374720} \end{APACrefDOI}
  \PrintBackRefs{\CurrentBib}
  
  \bibitem [\protect \citeauthoryear {%
    Der%
    , G{\"u}ttler%
    \BCBL {}\ \BBA {} Ay%
  }{%
    Der%
    \ \protect \BOthers {.}}{%
    {\protect \APACyear {2008}}%
  }]{%
    DerGuettlerEtAl-08}
  \APACinsertmetastar {%
    DerGuettlerEtAl-08}%
  \begin{APACrefauthors}%
    Der, R.%
    , G{\"u}ttler, F.%
    \BCBL {}\ \BBA {} Ay, N.%
  \end{APACrefauthors}%
  \unskip\
  \newblock
  \APACrefYearMonthDay{2008}{}{}.
  \newblock
  {\BBOQ}\APACrefatitle {{Predictive information and emergent cooperativity 
  in a
      chain of mobile robots}} {{Predictive information and emergent 
      cooperativity
      in a chain of mobile robots}}.{\BBCQ}
  \newblock
  \BIn{} S.~Bullock, J.~Noble, R.~Watson\BCBL {}\ \BBA {} M\BPBI A.~Bedau\
  (\BEDS), \APACrefbtitle {{The 11th International Conference on the 
  Synthesis
      and Simulation of Living Systems (Artificial Life XI)}} {{The 11th
      International Conference on the Synthesis and Simulation of Living 
      Systems
      (Artificial Life XI)}}\ (\BPGS\ 166--172).
  \newblock
  \APACaddressPublisher{}{{MIT Press}}.
  \newblock
  \begin{APACrefURL}
    \url{https://mitpress-request.mit.edu/sites/default/files/titles/alife/0262287196chap22.pdf}
  \end{APACrefURL}
  \PrintBackRefs{\CurrentBib}
  
  \bibitem [\protect \citeauthoryear {%
    Der%
    \ \BBA {} Martius%
  }{%
    Der%
    \ \BBA {} Martius%
  }{%
    {\protect \APACyear {2006}}%
  }]{%
    DerMartius-06}
  \APACinsertmetastar {%
    DerMartius-06}%
  \begin{APACrefauthors}%
    Der, R.%
    \BCBT {}\ \BBA {} Martius, G.%
  \end{APACrefauthors}%
  \unskip\
  \newblock
  \APACrefYearMonthDay{2006}{}{}.
  \newblock
  {\BBOQ}\APACrefatitle {{From Motor Babbling to Purposive Actions: Emerging
      Self-exploration in a Dynamical Systems Approach to Early Robot 
      Development}}
  {{From Motor Babbling to Purposive Actions: Emerging Self-exploration in a
      Dynamical Systems Approach to Early Robot Development}}.{\BBCQ}
  \newblock
  \BIn{} S.~Nolfi\ \BOthers {.}\ (\BEDS), \APACrefbtitle {{From Animals to
      Animats 9}} {{From Animals to Animats 9}}\ (\BVOL\ 4095, \BPGS\ 
      406--421).
  \newblock
  \APACaddressPublisher{}{Springer Berlin Heidelberg}.
  \newblock
  \begin{APACrefDOI} \doi{10.1007/11840541_34} \end{APACrefDOI}
  \PrintBackRefs{\CurrentBib}
  
  \bibitem [\protect \citeauthoryear {%
    Der%
    \ \BBA {} Martius%
  }{%
    Der%
    \ \BBA {} Martius%
  }{%
    {\protect \APACyear {2012}}%
  }]{%
    DerMartius-12}
  \APACinsertmetastar {%
    DerMartius-12}%
  \begin{APACrefauthors}%
    Der, R.%
    \BCBT {}\ \BBA {} Martius, G.%
  \end{APACrefauthors}%
  \unskip\
  \newblock
  \APACrefYear{2012}.
  \newblock
  \APACrefbtitle {{The Playful Machine}} {{The Playful Machine}}\ 
  (\BVOL~15).
  \newblock
  \APACaddressPublisher{}{Springer-Verlag Berlin Heidelberg}.
  \newblock
  \begin{APACrefDOI} \doi{10.1007/978-3-642-20253-7} \end{APACrefDOI}
  \PrintBackRefs{\CurrentBib}
  
  \bibitem [\protect \citeauthoryear {%
    Fiske%
  }{%
    Fiske%
  }{%
    {\protect \APACyear {2018}}%
  }]{%
    Fiske-18}
  \APACinsertmetastar {%
    Fiske-18}%
  \begin{APACrefauthors}%
    Fiske, S\BPBI T.%
  \end{APACrefauthors}%
  \unskip\
  \newblock
  \APACrefYearMonthDay{2018}{}{}.
  \newblock
  {\BBOQ}\APACrefatitle {{Stereotype Content: Warmth and Competence Endure}}
  {{Stereotype Content: Warmth and Competence Endure}}.{\BBCQ}
  \newblock
  \APACjournalVolNumPages{Current Directions in Psychological
    Science}{27}{2}{67--73}.
  \newblock
  \begin{APACrefDOI} \doi{10.1177/0963721417738825} \end{APACrefDOI}
  \PrintBackRefs{\CurrentBib}
  
  \bibitem [\protect \citeauthoryear {%
    Fiske%
    , Cuddy%
    \BCBL {}\ \BBA {} Glick%
  }{%
    Fiske%
    \ \protect \BOthers {.}}{%
    {\protect \APACyear {2007}}%
  }]{%
    FiskeCuddyEtAl-07}
  \APACinsertmetastar {%
    FiskeCuddyEtAl-07}%
  \begin{APACrefauthors}%
    Fiske, S\BPBI T.%
    , Cuddy, A\BPBI J\BPBI C.%
    \BCBL {}\ \BBA {} Glick, P.%
  \end{APACrefauthors}%
  \unskip\
  \newblock
  \APACrefYearMonthDay{2007}{}{}.
  \newblock
  {\BBOQ}\APACrefatitle {{Universal dimensions of social cognition: warmth 
  and
      competence}} {{Universal dimensions of social cognition: warmth and
      competence}}.{\BBCQ}
  \newblock
  \APACjournalVolNumPages{{Trends in Cognitive Sciences}}{11}{2}{77--83}.
  \newblock
  \begin{APACrefDOI} \doi{10.1016/j.tics.2006.11.005} \end{APACrefDOI}
  \PrintBackRefs{\CurrentBib}
  
  \bibitem [\protect \citeauthoryear {%
    Friston%
  }{%
    Friston%
  }{%
    {\protect \APACyear {2010}}%
  }]{%
    Friston-10}
  \APACinsertmetastar {%
    Friston-10}%
  \begin{APACrefauthors}%
    Friston, K.%
  \end{APACrefauthors}%
  \unskip\
  \newblock
  \APACrefYearMonthDay{2010}{}{}.
  \newblock
  {\BBOQ}\APACrefatitle {{The free-energy principle: a unified brain 
  theory?}}
  {{The free-energy principle: a unified brain theory?}}{\BBCQ}
  \newblock
  \APACjournalVolNumPages{Nature Reviews Neuroscience}{11}{}{127--138}.
  \newblock
  \begin{APACrefDOI} \doi{10.1038/nrn2787} \end{APACrefDOI}
  \PrintBackRefs{\CurrentBib}
  
  \bibitem [\protect \citeauthoryear {%
    Froese%
    \ \BBA {} Ziemke%
  }{%
    Froese%
    \ \BBA {} Ziemke%
  }{%
    {\protect \APACyear {2009}}%
  }]{%
    FroeseZiemke-09}
  \APACinsertmetastar {%
    FroeseZiemke-09}%
  \begin{APACrefauthors}%
    Froese, T.%
    \BCBT {}\ \BBA {} Ziemke, T.%
  \end{APACrefauthors}%
  \unskip\
  \newblock
  \APACrefYearMonthDay{2009}{}{}.
  \newblock
  {\BBOQ}\APACrefatitle {{Enactive artificial intelligence: Investigating 
  the
      systemic organization of life and mind}} {{Enactive artificial 
      intelligence:
      Investigating the systemic organization of life and mind}}.{\BBCQ}
  \newblock
  \APACjournalVolNumPages{Artificial Intelligence}{173}{3-4}{466--500}.
  \newblock
  \begin{APACrefDOI} \doi{10.1016/j.artint.2008.12.001} \end{APACrefDOI}
  \PrintBackRefs{\CurrentBib}
  
  \bibitem [\protect \citeauthoryear {%
    Fukuda%
    \ \BBA {} Ueda%
  }{%
    Fukuda%
    \ \BBA {} Ueda%
  }{%
    {\protect \APACyear {2010}}%
  }]{%
    FukudaUeda-10}
  \APACinsertmetastar {%
    FukudaUeda-10}%
  \begin{APACrefauthors}%
    Fukuda, H.%
    \BCBT {}\ \BBA {} Ueda, K.%
  \end{APACrefauthors}%
  \unskip\
  \newblock
  \APACrefYearMonthDay{2010}{}{}.
  \newblock
  {\BBOQ}\APACrefatitle {{Interaction with a Moving Object Affects One's
      Perception of Its Animacy}} {{Interaction with a Moving Object 
      Affects One's
      Perception of Its Animacy}}.{\BBCQ}
  \newblock
  \APACjournalVolNumPages{International Journal of Social
    Robotics}{2}{2}{187--193}.
  \newblock
  \begin{APACrefDOI} \doi{10.1007/s12369-010-0045-z} \end{APACrefDOI}
  \PrintBackRefs{\CurrentBib}
  
  \bibitem [\protect \citeauthoryear {%
    Geukes%
    \ \protect \BOthers {.}}{%
    Geukes%
    \ \protect \BOthers {.}}{%
    {\protect \APACyear {2019}}%
  }]{%
    GeukesBreilEtAl-19}
  \APACinsertmetastar {%
    GeukesBreilEtAl-19}%
  \begin{APACrefauthors}%
    Geukes, K.%
    , Breil, S\BPBI M.%
    , Hutteman, R.%
    , Nestler, S.%
    , Küfner, A\BPBI C\BPBI P.%
    \BCBL {}\ \BBA {} Back, M\BPBI D.%
  \end{APACrefauthors}%
  \unskip\
  \newblock
  \APACrefYearMonthDay{2019}{}{}.
  \newblock
  {\BBOQ}\APACrefatitle {{Explaining the longitudinal interplay of 
  personality
      and social relationships in the laboratory and in the field: The PILS 
      and the
      CONNECT study}} {{Explaining the longitudinal interplay of 
      personality and
      social relationships in the laboratory and in the field: The PILS and 
      the
      CONNECT study}}.{\BBCQ}
  \newblock
  \APACjournalVolNumPages{{PLOS} {ONE}}{14}{1}{}.
  \newblock
  \begin{APACrefDOI} \doi{10.1371/journal.pone.0210424} \end{APACrefDOI}
  \PrintBackRefs{\CurrentBib}
  
  \bibitem [\protect \citeauthoryear {%
    Gordon%
    , Breazeal%
    \BCBL {}\ \BBA {} Engel%
  }{%
    Gordon%
    \ \protect \BOthers {.}}{%
    {\protect \APACyear {2015}}%
  }]{%
    GordonBreazealEtAl-15}
  \APACinsertmetastar {%
    GordonBreazealEtAl-15}%
  \begin{APACrefauthors}%
    Gordon, G.%
    , Breazeal, C.%
    \BCBL {}\ \BBA {} Engel, S.%
  \end{APACrefauthors}%
  \unskip\
  \newblock
  \APACrefYearMonthDay{2015}{}{}.
  \newblock
  {\BBOQ}\APACrefatitle {{Can Children Catch Curiosity from a Social 
  Robot?}}
  {{Can Children Catch Curiosity from a Social Robot?}}{\BBCQ}
  \newblock
  \BIn{} \APACrefbtitle {{Proceedings of the Tenth Annual ACM/IEEE 
  International
      Conference on Human-Robot Interaction - HRI'15}.} {{Proceedings of 
      the Tenth
      Annual ACM/IEEE International Conference on Human-Robot Interaction -
      HRI'15}.}
  \newblock
  \APACaddressPublisher{}{{ACM} Press}.
  \newblock
  \begin{APACrefDOI} \doi{10.1145/2696454.2696469} \end{APACrefDOI}
  \PrintBackRefs{\CurrentBib}
  
  \bibitem [\protect \citeauthoryear {%
    Grassberger%
  }{%
    Grassberger%
  }{%
    {\protect \APACyear {1986}}%
  }]{%
    Grassberger-86}
  \APACinsertmetastar {%
    Grassberger-86}%
  \begin{APACrefauthors}%
    Grassberger, P.%
  \end{APACrefauthors}%
  \unskip\
  \newblock
  \APACrefYearMonthDay{1986}{}{}.
  \newblock
  {\BBOQ}\APACrefatitle {{Toward a quantitative theory of self-generated
      complexity}} {{Toward a quantitative theory of self-generated
      complexity}}.{\BBCQ}
  \newblock
  \APACjournalVolNumPages{International Journal of Theoretical
    Physics}{25}{9}{907--938}.
  \newblock
  \begin{APACrefDOI} \doi{10.1007/bf00668821} \end{APACrefDOI}
  \PrintBackRefs{\CurrentBib}
  
  \bibitem [\protect \citeauthoryear {%
    Guckelsberger%
    , Salge%
    \BCBL {}\ \BBA {} Colton%
  }{%
    Guckelsberger%
    \ \protect \BOthers {.}}{%
    {\protect \APACyear {2016}}%
  }]{%
    GuckelsbergerSalgeEtAl-16}
  \APACinsertmetastar {%
    GuckelsbergerSalgeEtAl-16}%
  \begin{APACrefauthors}%
    Guckelsberger, C.%
    , Salge, C.%
    \BCBL {}\ \BBA {} Colton, S.%
  \end{APACrefauthors}%
  \unskip\
  \newblock
  \APACrefYearMonthDay{2016}{}{}.
  \newblock
  {\BBOQ}\APACrefatitle {{Intrinsically motivated general companion NPCs via
      Coupled Empowerment Maximisation}} {{Intrinsically motivated general
      companion NPCs via Coupled Empowerment Maximisation}}.{\BBCQ}
  \newblock
  \BIn{} \APACrefbtitle {{2016 IEEE Conference on Computational 
  Intelligence and
      Games (CIG)}} {{2016 IEEE Conference on Computational Intelligence 
      and Games
      (CIG)}}\ (\BPGS\ 150--157).
  \newblock
  \APACaddressPublisher{}{{IEEE}}.
  \newblock
  \begin{APACrefDOI} \doi{10.1109/cig.2016.7860406} \end{APACrefDOI}
  \PrintBackRefs{\CurrentBib}
  
  \bibitem [\protect \citeauthoryear {%
    Guckelsberger%
    , Salge%
    \BCBL {}\ \BBA {} Togelius%
  }{%
    Guckelsberger%
    \ \protect \BOthers {.}}{%
    {\protect \APACyear {2018}}%
  }]{%
    GuckelsbergerSalgeEtAl-18}
  \APACinsertmetastar {%
    GuckelsbergerSalgeEtAl-18}%
  \begin{APACrefauthors}%
    Guckelsberger, C.%
    , Salge, C.%
    \BCBL {}\ \BBA {} Togelius, J.%
  \end{APACrefauthors}%
  \unskip\
  \newblock
  \APACrefYearMonthDay{2018}{}{}.
  \newblock
  {\BBOQ}\APACrefatitle {{New And Surprising Ways to Be Mean}} {{New And
      Surprising Ways to Be Mean}}.{\BBCQ}
  \newblock
  \BIn{} \APACrefbtitle {{2018 IEEE Conference on Computational 
  Intelligence and
      Games (CIG)}} {{2018 IEEE Conference on Computational Intelligence 
      and Games
      (CIG)}}\ (\BPGS\ 1--8).
  \newblock
  \APACaddressPublisher{}{{IEEE}}.
  \newblock
  \begin{APACrefDOI} \doi{10.1109/cig.2018.8490453} \end{APACrefDOI}
  \PrintBackRefs{\CurrentBib}
  
  \bibitem [\protect \citeauthoryear {%
    Hayashi%
    , Shiomi%
    , Kanda%
    \BCBL {}\ \BBA {} Hagita%
  }{%
    Hayashi%
    \ \protect \BOthers {.}}{%
    {\protect \APACyear {2010}}%
  }]{%
    HayashiShiomiEtAl-10}
  \APACinsertmetastar {%
    HayashiShiomiEtAl-10}%
  \begin{APACrefauthors}%
    Hayashi, K.%
    , Shiomi, M.%
    , Kanda, T.%
    \BCBL {}\ \BBA {} Hagita, N.%
  \end{APACrefauthors}%
  \unskip\
  \newblock
  \APACrefYearMonthDay{2010}{}{}.
  \newblock
  {\BBOQ}\APACrefatitle {{Who is appropriate? A robot, human and mascot 
  perform
      three troublesome tasks}} {{Who is appropriate? A robot, human and 
      mascot
      perform three troublesome tasks}}.{\BBCQ}
  \newblock
  \BIn{} \APACrefbtitle {{19th International Symposium in Robot and Human
      Interactive Communication}.} {{19th International Symposium in Robot 
      and
      Human Interactive Communication}.}
  \newblock
  \APACaddressPublisher{}{{IEEE}}.
  \newblock
  \begin{APACrefDOI} \doi{10.1109/roman.2010.5598661} \end{APACrefDOI}
  \PrintBackRefs{\CurrentBib}
  
  \bibitem [\protect \citeauthoryear {%
    Huang%
    \ \protect \BOthers {.}}{%
    Huang%
    \ \protect \BOthers {.}}{%
    {\protect \APACyear {2004}}%
  }]{%
    huang2004autonomy}
  \APACinsertmetastar {%
    huang2004autonomy}%
  \begin{APACrefauthors}%
    Huang, H\BHBI M.%
    , Messina, E.%
    , Wade, R.%
    , English, R.%
    , Novak, B.%
    \BCBL {}\ \BBA {} Albus, J.%
  \end{APACrefauthors}%
  \unskip\
  \newblock
  \APACrefYearMonthDay{2004}{}{}.
  \newblock
  {\BBOQ}\APACrefatitle {{Autonomy Measures for Robots}} {{Autonomy 
  Measures for
      Robots}}.{\BBCQ}
  \newblock
  \BIn{} \APACrefbtitle {{ASME 2004 International Mechanical Engineering 
  Congress
      and Exposition}} {{ASME 2004 International Mechanical Engineering 
      Congress
      and Exposition}}\ (\BPGS\ 1241--1247).
  \newblock
  \begin{APACrefDOI} \doi{10.1115/IMECE2004-61812} \end{APACrefDOI}
  \PrintBackRefs{\CurrentBib}
  
  \bibitem [\protect \citeauthoryear {%
    Jaques%
    \ \protect \BOthers {.}}{%
    Jaques%
    \ \protect \BOthers {.}}{%
    {\protect \APACyear {2019}}%
  }]{%
    JaquesLazaridouEtAl-19}
  \APACinsertmetastar {%
    JaquesLazaridouEtAl-19}%
  \begin{APACrefauthors}%
    Jaques, N.%
    , Lazaridou, A.%
    , Hughes, E.%
    , Gulcehre, C.%
    , Ortega, P.%
    , Strouse, D.%
    \BDBL {}De~Freitas, N.%
  \end{APACrefauthors}%
  \unskip\
  \newblock
  \APACrefYearMonthDay{2019}{}{}.
  \newblock
  {\BBOQ}\APACrefatitle {Social Influence as Intrinsic Motivation for 
  Multi-Agent
    Deep Reinforcement Learning} {Social influence as intrinsic motivation 
    for
    multi-agent deep reinforcement learning}.{\BBCQ}
  \newblock
  \BIn{} K.~Chaudhuri\ \BBA {} R.~Salakhutdinov\ (\BEDS), \APACrefbtitle
  {Proceedings of the 36th International Conference on Machine Learning}
  {Proceedings of the 36th international conference on machine learning}\
  (\BVOL~97, \BPGS\ 3040--3049).
  \newblock
  \APACaddressPublisher{}{PMLR}.
  \newblock
  \begin{APACrefURL} \url{https://proceedings.mlr.press/v97/jaques19a.html}
  \end{APACrefURL}
  \PrintBackRefs{\CurrentBib}
  
  \bibitem [\protect \citeauthoryear {%
    Judd%
    , James-Hawkins%
    , Yzerbyt%
    \BCBL {}\ \BBA {} Kashima%
  }{%
    Judd%
    \ \protect \BOthers {.}}{%
    {\protect \APACyear {2005}}%
  }]{%
    JuddJames-HawkinsEtAl-05}
  \APACinsertmetastar {%
    JuddJames-HawkinsEtAl-05}%
  \begin{APACrefauthors}%
    Judd, C\BPBI M.%
    , James-Hawkins, L.%
    , Yzerbyt, V.%
    \BCBL {}\ \BBA {} Kashima, Y.%
  \end{APACrefauthors}%
  \unskip\
  \newblock
  \APACrefYearMonthDay{2005}{}{}.
  \newblock
  {\BBOQ}\APACrefatitle {{Fundamental dimensions of social judgment:
      Understanding the relations between judgments of competence and 
      warmth}}
  {{Fundamental dimensions of social judgment: Understanding the relations
      between judgments of competence and warmth}}.{\BBCQ}
  \newblock
  \APACjournalVolNumPages{Journal of Personality and Social
    Psychology}{89}{6}{899--913}.
  \newblock
  \begin{APACrefDOI} \doi{10.1037/0022-3514.89.6.899} \end{APACrefDOI}
  \PrintBackRefs{\CurrentBib}
  
  \bibitem [\protect \citeauthoryear {%
    Kanda%
    , Shiomi%
    , Miyashita%
    , Ishiguro%
    \BCBL {}\ \BBA {} Hagita%
  }{%
    Kanda%
    \ \protect \BOthers {.}}{%
    {\protect \APACyear {2010}}%
  }]{%
    KandaShiomiEtAl-10}
  \APACinsertmetastar {%
    KandaShiomiEtAl-10}%
  \begin{APACrefauthors}%
    Kanda, T.%
    , Shiomi, M.%
    , Miyashita, Z.%
    , Ishiguro, H.%
    \BCBL {}\ \BBA {} Hagita, N.%
  \end{APACrefauthors}%
  \unskip\
  \newblock
  \APACrefYearMonthDay{2010}{}{}.
  \newblock
  {\BBOQ}\APACrefatitle {{A Communication Robot in a Shopping Mall}} {{A
      Communication Robot in a Shopping Mall}}.{\BBCQ}
  \newblock
  \APACjournalVolNumPages{{IEEE} Transactions on Robotics}{26}{5}{897--913}.
  \newblock
  \begin{APACrefDOI} \doi{10.1109/tro.2010.2062550} \end{APACrefDOI}
  \PrintBackRefs{\CurrentBib}
  
  \bibitem [\protect \citeauthoryear {%
    Kaplan%
    \ \BBA {} Oudeyer%
  }{%
    Kaplan%
    \ \BBA {} Oudeyer%
  }{%
    {\protect \APACyear {2004}}%
  }]{%
    KaplanOudeyer-04}
  \APACinsertmetastar {%
    KaplanOudeyer-04}%
  \begin{APACrefauthors}%
    Kaplan, F.%
    \BCBT {}\ \BBA {} Oudeyer, P\BHBI Y.%
  \end{APACrefauthors}%
  \unskip\
  \newblock
  \APACrefYearMonthDay{2004}{}{}.
  \newblock
  {\BBOQ}\APACrefatitle {{Maximizing Learning Progress: An Internal Reward 
  System
      for Development}} {{Maximizing Learning Progress: An Internal Reward 
      System
      for Development}}.{\BBCQ}
  \newblock
  \BIn{} F.~Iida, R.~Pfeifer, L.~Steels\BCBL {}\ \BBA {} Y.~Kuniyoshi\ 
  (\BEDS),
  \APACrefbtitle {{Embodied Artificial Intelligence}} {{Embodied Artificial
      Intelligence}}\ (\BVOL\ 3139, \BPGS\ 259--270).
  \newblock
  \APACaddressPublisher{}{Springer Berlin Heidelberg}.
  \newblock
  \begin{APACrefDOI} \doi{10.1007/978-3-540-27833-7_19} \end{APACrefDOI}
  \PrintBackRefs{\CurrentBib}
  
  \bibitem [\protect \citeauthoryear {%
    Klyubin%
    , Polani%
    \BCBL {}\ \BBA {} Nehaniv%
  }{%
    Klyubin%
    \ \protect \BOthers {.}}{%
    {\protect \APACyear {2005}}%
  }]{%
    KlyubinPolaniEtAl-05}
  \APACinsertmetastar {%
    KlyubinPolaniEtAl-05}%
  \begin{APACrefauthors}%
    Klyubin, A\BPBI S.%
    , Polani, D.%
    \BCBL {}\ \BBA {} Nehaniv, C\BPBI L.%
  \end{APACrefauthors}%
  \unskip\
  \newblock
  \APACrefYearMonthDay{2005}{}{}.
  \newblock
  {\BBOQ}\APACrefatitle {{Empowerment: A Universal Agent-Centric Measure of
      Control}} {{Empowerment: A Universal Agent-Centric Measure of
      Control}}.{\BBCQ}
  \newblock
  \BIn{} \APACrefbtitle {{2005 IEEE Congress on Evolutionary Computation}} 
  {{2005
      IEEE Congress on Evolutionary Computation}}\ (\BVOL~1, \BPGS\ 
      128--135).
  \newblock
  \APACaddressPublisher{}{{IEEE}}.
  \newblock
  \begin{APACrefDOI} \doi{10.1109/cec.2005.1554676} \end{APACrefDOI}
  \PrintBackRefs{\CurrentBib}
  
  \bibitem [\protect \citeauthoryear {%
    Kulms%
    \ \BBA {} Kopp%
  }{%
    Kulms%
    \ \BBA {} Kopp%
  }{%
    {\protect \APACyear {2018}}%
  }]{%
    KulmsKopp-18}
  \APACinsertmetastar {%
    KulmsKopp-18}%
  \begin{APACrefauthors}%
    Kulms, P.%
    \BCBT {}\ \BBA {} Kopp, S.%
  \end{APACrefauthors}%
  \unskip\
  \newblock
  \APACrefYearMonthDay{2018}{}{}.
  \newblock
  {\BBOQ}\APACrefatitle {{A Social Cognition Perspective on Human--Computer
      Trust: The Effect of Perceived Warmth and Competence on Trust in
      Decision-Making With Computers}} {{A Social Cognition Perspective on
      Human--Computer Trust: The Effect of Perceived Warmth and Competence 
      on Trust
      in Decision-Making With Computers}}.{\BBCQ}
  \newblock
  \APACjournalVolNumPages{Frontiers in Digital Humanities}{5}{}{}.
  \newblock
  \begin{APACrefDOI} \doi{10.3389/fdigh.2018.00014} \end{APACrefDOI}
  \PrintBackRefs{\CurrentBib}
  
  \bibitem [\protect \citeauthoryear {%
    {Lucasfilm Ltd.}%
  }{%
    {Lucasfilm Ltd.}%
  }{%
    {\protect \APACyear {2015}}%
  }]{%
    LucasfilmTwentiethCenturyFoxHomeEntertainment-15}
  \APACinsertmetastar {%
    LucasfilmTwentiethCenturyFoxHomeEntertainment-15}%
  \begin{APACrefauthors}%
    {Lucasfilm Ltd.}%
  \end{APACrefauthors}%
  \unskip\
  \newblock
  \APACrefYearMonthDay{2015}{}{}.
  \newblock
  \APACrefbtitle {{Star Wars}.} {{Star Wars}.}
  \newblock
  \APACrefnote{\url{https://www.starwars.com/films/}, accessed November 
  2020}
  \PrintBackRefs{\CurrentBib}
  
  \bibitem [\protect \citeauthoryear {%
    Martius%
  }{%
    Martius%
  }{%
    {\protect \APACyear {2013}}%
  }]{%
    tipi-implementation}
  \APACinsertmetastar {%
    tipi-implementation}%
  \begin{APACrefauthors}%
    Martius, G.%
  \end{APACrefauthors}%
  \unskip\
  \newblock
  \APACrefYearMonthDay{2013}{}{}.
  \newblock
  \APACrefbtitle {Implementation of predictive information maximization.}
  {Implementation of predictive information maximization.}
  \newblock
  \APACrefnote{\url{https://github.com/georgmartius/lpzrobots/blob/d2e6bbd164d902cdaa57eef154ed353ee0027236/selforg/controller/pimax.cpp},
    accessed December 2020}
  \PrintBackRefs{\CurrentBib}
  
  \bibitem [\protect \citeauthoryear {%
    Martius%
    , Der%
    \BCBL {}\ \BBA {} Ay%
  }{%
    Martius%
    \ \protect \BOthers {.}}{%
    {\protect \APACyear {2013}}%
    {\protect \APACexlab {{\protect \BCnt {1}}}}}]{%
    MartiusDerEtAl-13b}
  \APACinsertmetastar {%
    MartiusDerEtAl-13b}%
  \begin{APACrefauthors}%
    Martius, G.%
    , Der, R.%
    \BCBL {}\ \BBA {} Ay, N.%
  \end{APACrefauthors}%
  \unskip\
  \newblock
  \APACrefYearMonthDay{2013{\protect \BCnt {1}}}{}{}.
  \newblock
  \APACrefbtitle {{Appendix with derivations and technical detail}.} 
  {{Appendix
      with derivations and technical detail}.}
  \newblock
  \begin{APACrefDOI} \doi{https://doi.org/10.1371/journal.pone.0063400.s001}
  \end{APACrefDOI}
  \PrintBackRefs{\CurrentBib}
  
  \bibitem [\protect \citeauthoryear {%
    Martius%
    , Der%
    \BCBL {}\ \BBA {} Ay%
  }{%
    Martius%
    \ \protect \BOthers {.}}{%
    {\protect \APACyear {2013}}%
    {\protect \APACexlab {{\protect \BCnt {2}}}}}]{%
    MartiusDerEtAl-13}
  \APACinsertmetastar {%
    MartiusDerEtAl-13}%
  \begin{APACrefauthors}%
    Martius, G.%
    , Der, R.%
    \BCBL {}\ \BBA {} Ay, N.%
  \end{APACrefauthors}%
  \unskip\
  \newblock
  \APACrefYearMonthDay{2013{\protect \BCnt {2}}}{}{}.
  \newblock
  {\BBOQ}\APACrefatitle {{Information Driven Self-Organization of Complex 
  Robotic
      Behaviors}} {{Information Driven Self-Organization of Complex Robotic
      Behaviors}}.{\BBCQ}
  \newblock
  \APACjournalVolNumPages{PLOS One}{8}{5}{1--14}.
  \newblock
  \begin{APACrefDOI} \doi{10.1371/journal.pone.0063400} \end{APACrefDOI}
  \PrintBackRefs{\CurrentBib}
  
  \bibitem [\protect \citeauthoryear {%
    Martius%
    , Jahn%
    , Hauser%
    \BCBL {}\ \BBA {} Hafner%
  }{%
    Martius%
    \ \protect \BOthers {.}}{%
    {\protect \APACyear {2014}}%
  }]{%
    MartiusJahnEtAl-14}
  \APACinsertmetastar {%
    MartiusJahnEtAl-14}%
  \begin{APACrefauthors}%
    Martius, G.%
    , Jahn, L.%
    , Hauser, H.%
    \BCBL {}\ \BBA {} Hafner, V\BPBI V.%
  \end{APACrefauthors}%
  \unskip\
  \newblock
  \APACrefYearMonthDay{2014}{}{}.
  \newblock
  {\BBOQ}\APACrefatitle {{Self-exploration of the Stumpy Robot with 
  Predictive
      Information Maximization}} {{Self-exploration of the Stumpy Robot with
      Predictive Information Maximization}}.{\BBCQ}
  \newblock
  \BIn{} \APACrefbtitle {{From Animals to Animats 13}} {{From Animals to 
  Animats
      13}}\ (\BPGS\ 32--42).
  \newblock
  \APACaddressPublisher{}{Springer International Publishing}.
  \newblock
  \begin{APACrefDOI} \doi{10.1007/978-3-319-08864-8_4} \end{APACrefDOI}
  \PrintBackRefs{\CurrentBib}
  
  \bibitem [\protect \citeauthoryear {%
    Maturana%
    \ \BBA {} Varela%
  }{%
    Maturana%
    \ \BBA {} Varela%
  }{%
    {\protect \APACyear {1991}}%
  }]{%
    MaturanaVarela-91}
  \APACinsertmetastar {%
    MaturanaVarela-91}%
  \begin{APACrefauthors}%
    Maturana, H\BPBI R.%
    \BCBT {}\ \BBA {} Varela, F\BPBI J.%
  \end{APACrefauthors}%
  \unskip\
  \newblock
  \APACrefYear{1991}.
  \newblock
  \APACrefbtitle {{Autopoiesis and Cognition: The Realization of the 
  Living}}
  {{Autopoiesis and Cognition: The Realization of the Living}}\ (\BVOL~42;
  R\BPBI S.~Cohen\ \BBA {} M\BPBI W.~Wartofsky, \BEDS{}).
  \newblock
  \APACaddressPublisher{}{Springer Science \& Business Media}.
  \newblock
  \begin{APACrefDOI} \doi{10.1007/978-94-009-8947-4} \end{APACrefDOI}
  \PrintBackRefs{\CurrentBib}
  
  \bibitem [\protect \citeauthoryear {%
    Merrick%
    \ \BBA {} Maher%
  }{%
    Merrick%
    \ \BBA {} Maher%
  }{%
    {\protect \APACyear {2009}}%
  }]{%
    MerrickMaher-09}
  \APACinsertmetastar {%
    MerrickMaher-09}%
  \begin{APACrefauthors}%
    Merrick, K\BPBI E.%
    \BCBT {}\ \BBA {} Maher, M\BPBI L.%
  \end{APACrefauthors}%
  \unskip\
  \newblock
  \APACrefYear{2009}.
  \newblock
  \APACrefbtitle {{Motivated Reinforcement Learning}} {{Motivated 
  Reinforcement
      Learning}}\ (\PrintOrdinal{1}\ \BEd).
  \newblock
  \APACaddressPublisher{}{Springer Berlin Heidelberg}.
  \newblock
  \begin{APACrefDOI} \doi{10.1007/978-3-540-89187-1} \end{APACrefDOI}
  \PrintBackRefs{\CurrentBib}
  
  \bibitem [\protect \citeauthoryear {%
    Noguchi%
    , Gel%
    , Brunner%
    \BCBL {}\ \BBA {} Konietschke%
  }{%
    Noguchi%
    \ \protect \BOthers {.}}{%
    {\protect \APACyear {2012}}%
  }]{%
    NoguchiGelEtAl-12}
  \APACinsertmetastar {%
    NoguchiGelEtAl-12}%
  \begin{APACrefauthors}%
    Noguchi, K.%
    , Gel, Y\BPBI R.%
    , Brunner, E.%
    \BCBL {}\ \BBA {} Konietschke, F.%
  \end{APACrefauthors}%
  \unskip\
  \newblock
  \APACrefYearMonthDay{2012}{}{}.
  \newblock
  {\BBOQ}\APACrefatitle {{nparLD}: {AnRSoftware} Package for the 
  Nonparametric
    Analysis of Longitudinal Data in Factorial Experiments} {{nparLD}:
    {AnRSoftware} package for the nonparametric analysis of longitudinal 
    data in
    factorial experiments}.{\BBCQ}
  \newblock
  \APACjournalVolNumPages{Journal of Statistical Software}{50}{12}{}.
  \newblock
  \begin{APACrefDOI} \doi{10.18637/jss.v050.i12} \end{APACrefDOI}
  \PrintBackRefs{\CurrentBib}
  
  \bibitem [\protect \citeauthoryear {%
    Oliveira%
    , Arriaga%
    , Correia%
    \BCBL {}\ \BBA {} Paiva%
  }{%
    Oliveira%
    \ \protect \BOthers {.}}{%
    {\protect \APACyear {2019}}%
  }]{%
    OliveiraArriagaEtAl-19}
  \APACinsertmetastar {%
    OliveiraArriagaEtAl-19}%
  \begin{APACrefauthors}%
    Oliveira, R.%
    , Arriaga, P.%
    , Correia, F.%
    \BCBL {}\ \BBA {} Paiva, A.%
  \end{APACrefauthors}%
  \unskip\
  \newblock
  \APACrefYearMonthDay{2019}{}{}.
  \newblock
  {\BBOQ}\APACrefatitle {The Stereotype Content Model Applied to Human-Robot
    Interactions in Groups} {The stereotype content model applied to 
    human-robot
    interactions in groups}.{\BBCQ}
  \newblock
  \BIn{} \APACrefbtitle {2019 14th {ACM}/{IEEE} International Conference on
    Human-Robot Interaction ({HRI}).} {2019 14th {ACM}/{IEEE} international
    conference on human-robot interaction ({HRI}).}
  \newblock
  \APACaddressPublisher{}{{IEEE}}.
  \newblock
  \begin{APACrefDOI} \doi{10.1109/hri.2019.8673171} \end{APACrefDOI}
  \PrintBackRefs{\CurrentBib}
  
  \bibitem [\protect \citeauthoryear {%
    O'Regan%
    \ \BBA {} No{\"e}%
  }{%
    O'Regan%
    \ \BBA {} No{\"e}%
  }{%
    {\protect \APACyear {2001}}%
  }]{%
    o2001sensorimotor}
  \APACinsertmetastar {%
    o2001sensorimotor}%
  \begin{APACrefauthors}%
    O'Regan, J\BPBI K.%
    \BCBT {}\ \BBA {} No{\"e}, A.%
  \end{APACrefauthors}%
  \unskip\
  \newblock
  \APACrefYearMonthDay{2001}{}{}.
  \newblock
  {\BBOQ}\APACrefatitle {{A sensorimotor account of vision and visual
      consciousness}} {{A sensorimotor account of vision and visual
      consciousness}}.{\BBCQ}
  \newblock
  \APACjournalVolNumPages{Behavioral and brain sciences}{24}{5}{939--973}.
  \newblock
  \begin{APACrefDOI} \doi{10.1017/S0140525X01000115} \end{APACrefDOI}
  \PrintBackRefs{\CurrentBib}
  
  \bibitem [\protect \citeauthoryear {%
    Orne%
  }{%
    Orne%
  }{%
    {\protect \APACyear {1962}}%
  }]{%
    Orne-62}
  \APACinsertmetastar {%
    Orne-62}%
  \begin{APACrefauthors}%
    Orne, M\BPBI T.%
  \end{APACrefauthors}%
  \unskip\
  \newblock
  \APACrefYearMonthDay{1962}{}{}.
  \newblock
  {\BBOQ}\APACrefatitle {On the social psychology of the psychological
    experiment: With particular reference to demand characteristics and 
    their
    implications.} {On the social psychology of the psychological 
    experiment:
    With particular reference to demand characteristics and their
    implications.}{\BBCQ}
  \newblock
  \APACjournalVolNumPages{American Psychologist}{17}{11}{776--783}.
  \newblock
  \begin{APACrefDOI} \doi{10.1037/h0043424} \end{APACrefDOI}
  \PrintBackRefs{\CurrentBib}
  
  \bibitem [\protect \citeauthoryear {%
    Oudeyer%
    , Gottlieb%
    \BCBL {}\ \BBA {} Lopes%
  }{%
    Oudeyer%
    \ \protect \BOthers {.}}{%
    {\protect \APACyear {2016}}%
  }]{%
    OudeyerGottliebEtAl-16}
  \APACinsertmetastar {%
    OudeyerGottliebEtAl-16}%
  \begin{APACrefauthors}%
    Oudeyer, P\BHBI Y.%
    , Gottlieb, J.%
    \BCBL {}\ \BBA {} Lopes, M.%
  \end{APACrefauthors}%
  \unskip\
  \newblock
  \APACrefYearMonthDay{2016}{}{}.
  \newblock
  {\BBOQ}\APACrefatitle {{Intrinsic motivation, curiosity, and learning}}
  {{Intrinsic motivation, curiosity, and learning}}.{\BBCQ}
  \newblock
  \BIn{} \APACrefbtitle {Motivation - Theory, Neurobiology and Applications}
  {Motivation - theory, neurobiology and applications}\ (\BPGS\ 257--284).
  \newblock
  \APACaddressPublisher{}{Elsevier}.
  \newblock
  \begin{APACrefDOI} \doi{10.1016/bs.pbr.2016.05.005} \end{APACrefDOI}
  \PrintBackRefs{\CurrentBib}
  
  \bibitem [\protect \citeauthoryear {%
    Oudeyer%
    \ \BBA {} Kaplan%
  }{%
    Oudeyer%
    \ \BBA {} Kaplan%
  }{%
    {\protect \APACyear {2008}}%
  }]{%
    OudeyerKaplan-08}
  \APACinsertmetastar {%
    OudeyerKaplan-08}%
  \begin{APACrefauthors}%
    Oudeyer, P\BHBI Y.%
    \BCBT {}\ \BBA {} Kaplan, F.%
  \end{APACrefauthors}%
  \unskip\
  \newblock
  \APACrefYearMonthDay{2008}{}{}.
  \newblock
  {\BBOQ}\APACrefatitle {{How can we define intrinsic motivation ?}} {{How 
  can we
      define intrinsic motivation ?}}{\BBCQ}
  \newblock
  \BIn{} \APACrefbtitle {Proceedings of the 8th International Conference on
    Epigenetic Robotics: Modeling Cognitive Development in Robotic Systems.}
  {Proceedings of the 8th international conference on epigenetic robotics:
    Modeling cognitive development in robotic systems.}
  \newblock
  \APACaddressPublisher{Brighton, United Kingdom}{}.
  \newblock
  \begin{APACrefURL} \url{https://hal.inria.fr/inria-00420175} 
  \end{APACrefURL}
  \PrintBackRefs{\CurrentBib}
  
  \bibitem [\protect \citeauthoryear {%
    Oudeyer%
    \ \BBA {} Kaplan%
  }{%
    Oudeyer%
    \ \BBA {} Kaplan%
  }{%
    {\protect \APACyear {2009}}%
  }]{%
    OudeyerKaplan-09}
  \APACinsertmetastar {%
    OudeyerKaplan-09}%
  \begin{APACrefauthors}%
    Oudeyer, P\BHBI Y.%
    \BCBT {}\ \BBA {} Kaplan, F.%
  \end{APACrefauthors}%
  \unskip\
  \newblock
  \APACrefYearMonthDay{2009}{}{}.
  \newblock
  {\BBOQ}\APACrefatitle {{What is Intrinsic Motivation? A Typology of
      Computational Approaches}} {{What is Intrinsic Motivation? A Typology 
      of
      Computational Approaches}}.{\BBCQ}
  \newblock
  \APACjournalVolNumPages{Frontiers in Neurorobotics}{1}{}{6}.
  \newblock
  \begin{APACrefDOI} \doi{10.3389/neuro.12.006.2007} \end{APACrefDOI}
  \PrintBackRefs{\CurrentBib}
  
  \bibitem [\protect \citeauthoryear {%
    Oudeyer%
    , Kaplan%
    \BCBL {}\ \BBA {} Hafner%
  }{%
    Oudeyer%
    \ \protect \BOthers {.}}{%
    {\protect \APACyear {2007}}%
  }]{%
    OudeyerKaplanEtAl-07}
  \APACinsertmetastar {%
    OudeyerKaplanEtAl-07}%
  \begin{APACrefauthors}%
    Oudeyer, P\BHBI Y.%
    , Kaplan, F.%
    \BCBL {}\ \BBA {} Hafner, V\BPBI V.%
  \end{APACrefauthors}%
  \unskip\
  \newblock
  \APACrefYearMonthDay{2007}{}{}.
  \newblock
  {\BBOQ}\APACrefatitle {{Intrinsic Motivation Systems for Autonomous Mental
      Development}} {{Intrinsic Motivation Systems for Autonomous Mental
      Development}}.{\BBCQ}
  \newblock
  \APACjournalVolNumPages{{IEEE} Transactions on Evolutionary
    Computation}{11}{2}{265--286}.
  \newblock
  \begin{APACrefDOI} \doi{10.1109/tevc.2006.890271} \end{APACrefDOI}
  \PrintBackRefs{\CurrentBib}
  
  \bibitem [\protect \citeauthoryear {%
    Paolo%
  }{%
    Paolo%
  }{%
    {\protect \APACyear {2004}}%
  }]{%
    Paolo-04}
  \APACinsertmetastar {%
    Paolo-04}%
  \begin{APACrefauthors}%
    Paolo, E\BPBI A\BPBI D.%
  \end{APACrefauthors}%
  \unskip\
  \newblock
  \APACrefYear{2004}.
  \newblock
  \APACrefbtitle {{Unbinding Biological Autonomy: Francisco Varela's
      Contributions to Artificial Life}} {{Unbinding Biological Autonomy: 
      Francisco
      Varela's Contributions to Artificial Life}}\ (\BVOL~10)\ (\BNUM~3).
  \newblock
  \APACaddressPublisher{}{{MIT} Press - Journals}.
  \newblock
  \begin{APACrefDOI} \doi{10.1162/1064546041255566} \end{APACrefDOI}
  \PrintBackRefs{\CurrentBib}
  
  \bibitem [\protect \citeauthoryear {%
    Pinillos%
    , Marcos%
    , Feliz%
    , Zalama%
    \BCBL {}\ \BBA {} G{\'{o}}mez-Garc{\'{\i}}a-Bermejo%
  }{%
    Pinillos%
    \ \protect \BOthers {.}}{%
    {\protect \APACyear {2016}}%
  }]{%
    PinillosMarcosEtAl-16}
  \APACinsertmetastar {%
    PinillosMarcosEtAl-16}%
  \begin{APACrefauthors}%
    Pinillos, R.%
    , Marcos, S.%
    , Feliz, R.%
    , Zalama, E.%
    \BCBL {}\ \BBA {} G{\'{o}}mez-Garc{\'{\i}}a-Bermejo, J.%
  \end{APACrefauthors}%
  \unskip\
  \newblock
  \APACrefYearMonthDay{2016}{}{}.
  \newblock
  {\BBOQ}\APACrefatitle {{Long-term assessment of a service robot in a hotel
      environment}} {{Long-term assessment of a service robot in a hotel
      environment}}.{\BBCQ}
  \newblock
  \APACjournalVolNumPages{Robotics and Autonomous Systems}{79}{}{40--57}.
  \newblock
  \begin{APACrefDOI} \doi{10.1016/j.robot.2016.01.014} \end{APACrefDOI}
  \PrintBackRefs{\CurrentBib}
  
  \bibitem [\protect \citeauthoryear {%
    {Research Network for Self-Organization of Robot Behavior}%
  }{%
    {Research Network for Self-Organization of Robot Behavior}%
  }{%
    {\protect \APACyear {2015}}%
  }]{%
    playful-videos-20}
  \APACinsertmetastar {%
    playful-videos-20}%
  \begin{APACrefauthors}%
    {Research Network for Self-Organization of Robot Behavior}.%
  \end{APACrefauthors}%
  \unskip\
  \newblock
  \APACrefYearMonthDay{2015}{}{}.
  \newblock
  \APACrefbtitle {{Research Network for Self-Organization of Robot Behavior:
      Videos}.} {{Research Network for Self-Organization of Robot Behavior:
      Videos}.}
  \newblock
  \APACrefnote{\url{https://robot.informatik.uni-leipzig.de/videos}, 
  accessed
    February 2020}
  \PrintBackRefs{\CurrentBib}
  
  \bibitem [\protect \citeauthoryear {%
    Rich%
    , Ponsler%
    , Holroyd%
    \BCBL {}\ \BBA {} Sidner%
  }{%
    Rich%
    \ \protect \BOthers {.}}{%
    {\protect \APACyear {2010}}%
  }]{%
    RichPonslerEtAl-10}
  \APACinsertmetastar {%
    RichPonslerEtAl-10}%
  \begin{APACrefauthors}%
    Rich, C.%
    , Ponsler, B.%
    , Holroyd, A.%
    \BCBL {}\ \BBA {} Sidner, C\BPBI L.%
  \end{APACrefauthors}%
  \unskip\
  \newblock
  \APACrefYearMonthDay{2010}{}{}.
  \newblock
  {\BBOQ}\APACrefatitle {{Recognizing engagement in human-robot 
  interaction}}
  {{Recognizing engagement in human-robot interaction}}.{\BBCQ}
  \newblock
  \BIn{} \APACrefbtitle {{2010 5th ACM/IEEE International Conference on
      Human-Robot Interaction (HRI)}} {{2010 5th ACM/IEEE International 
      Conference
      on Human-Robot Interaction (HRI)}}\ (\BPGS\ 375--382).
  \newblock
  \begin{APACrefDOI} \doi{10.1109/HRI.2010.5453163} \end{APACrefDOI}
  \PrintBackRefs{\CurrentBib}
  
  \bibitem [\protect \citeauthoryear {%
    Rosas%
    \ \protect \BOthers {.}}{%
    Rosas%
    \ \protect \BOthers {.}}{%
    {\protect \APACyear {2020}}%
  }]{%
    RosasMedianoEtAl-20}
  \APACinsertmetastar {%
    RosasMedianoEtAl-20}%
  \begin{APACrefauthors}%
    Rosas, F\BPBI E.%
    , Mediano, P\BPBI A\BPBI M.%
    , Jensen, H\BPBI J.%
    , Seth, A\BPBI K.%
    , Barrett, A\BPBI B.%
    , Carhart-Harris, R\BPBI L.%
    \BCBL {}\ \BBA {} Bor, D.%
  \end{APACrefauthors}%
  \unskip\
  \newblock
  \APACrefYearMonthDay{2020}{}{}.
  \newblock
  {\BBOQ}\APACrefatitle {{Reconciling emergences: An information-theoretic
      approach to identify causal emergence in multivariate data}} 
      {{Reconciling
      emergences: An information-theoretic approach to identify causal 
      emergence in
      multivariate data}}.{\BBCQ}
  \newblock
  \APACjournalVolNumPages{{PLOS} Computational Biology}{16}{12}{e1008289}.
  \newblock
  \begin{APACrefDOI} \doi{10.1371/journal.pcbi.1008289} \end{APACrefDOI}
  \PrintBackRefs{\CurrentBib}
  
  \bibitem [\protect \citeauthoryear {%
    Rosenberg%
    , Nelson%
    \BCBL {}\ \BBA {} Vivekananthan%
  }{%
    Rosenberg%
    \ \protect \BOthers {.}}{%
    {\protect \APACyear {1968}}%
  }]{%
    RosenbergNelsonEtAl-68}
  \APACinsertmetastar {%
    RosenbergNelsonEtAl-68}%
  \begin{APACrefauthors}%
    Rosenberg, S.%
    , Nelson, C.%
    \BCBL {}\ \BBA {} Vivekananthan, P\BPBI S.%
  \end{APACrefauthors}%
  \unskip\
  \newblock
  \APACrefYearMonthDay{1968}{}{}.
  \newblock
  {\BBOQ}\APACrefatitle {A multidimensional approach to the structure of
    personality impressions.} {A multidimensional approach to the structure 
    of
    personality impressions.}{\BBCQ}
  \newblock
  \APACjournalVolNumPages{Journal of Personality and Social
    Psychology}{9}{4}{283--294}.
  \newblock
  \begin{APACrefDOI} \doi{10.1037/h0026086} \end{APACrefDOI}
  \PrintBackRefs{\CurrentBib}
  
  \bibitem [\protect \citeauthoryear {%
    Ryan%
    \ \BBA {} Deci%
  }{%
    Ryan%
    \ \BBA {} Deci%
  }{%
    {\protect \APACyear {2000}}%
    {\protect \APACexlab {{\protect \BCnt {1}}}}}]{%
    RyanDeci-00}
  \APACinsertmetastar {%
    RyanDeci-00}%
  \begin{APACrefauthors}%
    Ryan, R\BPBI M.%
    \BCBT {}\ \BBA {} Deci, E\BPBI L.%
  \end{APACrefauthors}%
  \unskip\
  \newblock
  \APACrefYearMonthDay{2000{\protect \BCnt {1}}}{}{}.
  \newblock
  {\BBOQ}\APACrefatitle {{Intrinsic and Extrinsic Motivations: Classic
      Definitions and New Directions}} {{Intrinsic and Extrinsic 
      Motivations:
      Classic Definitions and New Directions}}.{\BBCQ}
  \newblock
  \APACjournalVolNumPages{Contemporary Educational 
  Psychology}{25}{1}{54--67}.
  \newblock
  \begin{APACrefDOI} \doi{10.1006/ceps.1999.1020} \end{APACrefDOI}
  \PrintBackRefs{\CurrentBib}
  
  \bibitem [\protect \citeauthoryear {%
    Ryan%
    \ \BBA {} Deci%
  }{%
    Ryan%
    \ \BBA {} Deci%
  }{%
    {\protect \APACyear {2000}}%
    {\protect \APACexlab {{\protect \BCnt {2}}}}}]{%
    ryan2000self}
  \APACinsertmetastar {%
    ryan2000self}%
  \begin{APACrefauthors}%
    Ryan, R\BPBI M.%
    \BCBT {}\ \BBA {} Deci, E\BPBI L.%
  \end{APACrefauthors}%
  \unskip\
  \newblock
  \APACrefYearMonthDay{2000{\protect \BCnt {2}}}{}{}.
  \newblock
  {\BBOQ}\APACrefatitle {{Self-determination theory and the facilitation of
      intrinsic motivation, social development, and well-being}}
  {{Self-determination theory and the facilitation of intrinsic motivation,
      social development, and well-being}}.{\BBCQ}
  \newblock
  \APACjournalVolNumPages{American Psychologist}{55}{1}{68--78}.
  \newblock
  \begin{APACrefDOI} \doi{10.1037/0003-066X.55.1.68} \end{APACrefDOI}
  \PrintBackRefs{\CurrentBib}
  
  \bibitem [\protect \citeauthoryear {%
    Scheunemann%
  }{%
    Scheunemann%
  }{%
    {\protect \APACyear {2018}}%
  }]{%
    SPHEROCPPrep}
  \APACinsertmetastar {%
    SPHEROCPPrep}%
  \begin{APACrefauthors}%
    Scheunemann, M\BPBI M.%
  \end{APACrefauthors}%
  \unskip\
  \newblock
  \APACrefYearMonthDay{2018}{}{}.
  \newblock
  \APACrefbtitle {{Code repository: spherocpp}.} {{Code repository: 
  spherocpp}.}
  \newblock
  \begin{APACrefURL} 
  [{2020-12-12}]\url{https://gitlab.com/scheunemann/spheropp}
  \end{APACrefURL}
  \newblock
  \APACrefnote{\url{https://gitlab.com/scheunemann/spheropp}, accessed 
  December
    2020}
  \PrintBackRefs{\CurrentBib}
  
  \bibitem [\protect \citeauthoryear {%
    Scheunemann%
  }{%
    Scheunemann%
  }{%
    {\protect \APACyear {2021}}%
    {\protect \APACexlab {{\protect \BCnt {1}}}}}]{%
    Scheunemann-21}
  \APACinsertmetastar {%
    Scheunemann-21}%
  \begin{APACrefauthors}%
    Scheunemann, M\BPBI M.%
  \end{APACrefauthors}%
  \unskip\
  \newblock
  \APACrefYear{2021{\protect \BCnt {1}}}.
  \unskip\
  \newblock
  \APACrefbtitle {{Autonomous and Intrinsically Motivated Robots for 
  Sustained
      Human-Robot Interaction}} {{Autonomous and Intrinsically Motivated 
      Robots for
      Sustained Human-Robot Interaction}}\ \APACtypeAddressSchool 
      {\BPhD}{}{}.
  \unskip\
  \newblock
  \begin{APACrefDOI} \doi{10.18745/TH.23936} \end{APACrefDOI}
  \PrintBackRefs{\CurrentBib}
  
  \bibitem [\protect \citeauthoryear {%
    Scheunemann%
  }{%
    Scheunemann%
  }{%
    {\protect \APACyear {2021}}%
    {\protect \APACexlab {{\protect \BCnt {2}}}}}]{%
    supp-video}
  \APACinsertmetastar {%
    supp-video}%
  \begin{APACrefauthors}%
    Scheunemann, M\BPBI M.%
  \end{APACrefauthors}%
  \unskip\
  \newblock
  \APACrefYearMonthDay{2021{\protect \BCnt {2}}}{}{}.
  \newblock
  \APACrefbtitle {{Supplementary Material for Human Perception of 
  Intrinsically
      Motivated Autonomy in Human-Robot Interaction}.} {{Supplementary 
      Material for
      Human Perception of Intrinsically Motivated Autonomy in Human-Robot
      Interaction}.}
  \newblock
  \APACrefnote{\url{https://mms.ai/adb2021}, accessed November 2021}
  \PrintBackRefs{\CurrentBib}
  
  \bibitem [\protect \citeauthoryear {%
    Scheunemann%
    , Cuijpers%
    \BCBL {}\ \BBA {} Salge%
  }{%
    Scheunemann%
    \ \protect \BOthers {.}}{%
    {\protect \APACyear {2020}}%
  }]{%
    ScheunemannCuijpersEtAl-20}
  \APACinsertmetastar {%
    ScheunemannCuijpersEtAl-20}%
  \begin{APACrefauthors}%
    Scheunemann, M\BPBI M.%
    , Cuijpers, R\BPBI H.%
    \BCBL {}\ \BBA {} Salge, C.%
  \end{APACrefauthors}%
  \unskip\
  \newblock
  \APACrefYearMonthDay{2020}{}{}.
  \newblock
  {\BBOQ}\APACrefatitle {{Warmth and Competence to Predict Human Preference 
  of
      Robot Behavior in Physical Human-Robot Interaction}} {{Warmth and 
      Competence
      to Predict Human Preference of Robot Behavior in Physical Human-Robot
      Interaction}}.{\BBCQ}
  \newblock
  \BIn{} \APACrefbtitle {{Proceedings of the 29th IEEE International 
  Symposium on
      Robot and Human Interactive Communication (RO-MAN)}} {{Proceedings of 
      the
      29th IEEE International Symposium on Robot and Human Interactive
      Communication (RO-MAN)}}\ (\BPGS\ 1340--1347).
  \newblock
  \APACaddressPublisher{}{IEEE}.
  \newblock
  \begin{APACrefDOI} \doi{10.1109/RO-MAN47096.2020.9223478} \end{APACrefDOI}
  \PrintBackRefs{\CurrentBib}
  
  \bibitem [\protect \citeauthoryear {%
    Scheunemann%
    \ \BBA {} Dautenhahn%
  }{%
    Scheunemann%
    \ \BBA {} Dautenhahn%
  }{%
    {\protect \APACyear {2017}}%
  }]{%
    ScheunemannDautenhahn-17}
  \APACinsertmetastar {%
    ScheunemannDautenhahn-17}%
  \begin{APACrefauthors}%
    Scheunemann, M\BPBI M.%
    \BCBT {}\ \BBA {} Dautenhahn, K.%
  \end{APACrefauthors}%
  \unskip\
  \newblock
  \APACrefYearMonthDay{2017}{}{}.
  \newblock
  {\BBOQ}\APACrefatitle {Bluetooth Low Energy for Autonomous Human-Robot
    Interaction} {Bluetooth low energy for autonomous human-robot
    interaction}.{\BBCQ}
  \newblock
  \BIn{} \APACrefbtitle {Proceedings of the Companion of the 2017 
  {ACM}/{IEEE}
    International Conference on Human-Robot Interaction.} {Proceedings of 
    the
    companion of the 2017 {ACM}/{IEEE} international conference on 
    human-robot
    interaction.}
  \newblock
  \APACaddressPublisher{}{{ACM}}.
  \newblock
  \begin{APACrefDOI} \doi{10.1145/3029798.3036663} \end{APACrefDOI}
  \PrintBackRefs{\CurrentBib}
  
  \bibitem [\protect \citeauthoryear {%
    Scheunemann%
    , Salge%
    \BCBL {}\ \BBA {} Dautenhahn%
  }{%
    Scheunemann%
    \ \protect \BOthers {.}}{%
    {\protect \APACyear {2019}}%
  }]{%
    ScheunemannSalgeEtAl-19}
  \APACinsertmetastar {%
    ScheunemannSalgeEtAl-19}%
  \begin{APACrefauthors}%
    Scheunemann, M\BPBI M.%
    , Salge, C.%
    \BCBL {}\ \BBA {} Dautenhahn, K.%
  \end{APACrefauthors}%
  \unskip\
  \newblock
  \APACrefYearMonthDay{2019}{}{}.
  \newblock
  {\BBOQ}\APACrefatitle {{Intrinsically Motivated Autonomy in Human-Robot
      Interaction: Human Perception of Predictive Information in Robots}}
  {{Intrinsically Motivated Autonomy in Human-Robot Interaction: Human
      Perception of Predictive Information in Robots}}.{\BBCQ}
  \newblock
  \BIn{} K.~Althoefer, J.~Konstantinova\BCBL {}\ \BBA {} K.~Zhang\ (\BEDS),
  \APACrefbtitle {Towards Autonomous Robotic Systems} {Towards autonomous
    robotic systems}\ (\BPGS\ 325--337).
  \newblock
  \APACaddressPublisher{Cham}{Springer International Publishing}.
  \newblock
  \begin{APACrefDOI} \doi{10.1007/978-3-030-23807-0_27} \end{APACrefDOI}
  \PrintBackRefs{\CurrentBib}
  
  \bibitem [\protect \citeauthoryear {%
    Schmidhuber%
  }{%
    Schmidhuber%
  }{%
    {\protect \APACyear {1991}}%
  }]{%
    Schmidhuber-91}
  \APACinsertmetastar {%
    Schmidhuber-91}%
  \begin{APACrefauthors}%
    Schmidhuber, J.%
  \end{APACrefauthors}%
  \unskip\
  \newblock
  \APACrefYearMonthDay{1991}{}{}.
  \newblock
  {\BBOQ}\APACrefatitle {{Curious model-building control systems}} {{Curious
      model-building control systems}}.{\BBCQ}
  \newblock
  \BIn{} \APACrefbtitle {[Proceedings] 1991 IEEE International Joint 
  Conference
    on Neural Networks} {[proceedings] 1991 ieee international joint 
    conference
    on neural networks}\ (\BPGS\ 1458--1463).
  \newblock
  \APACaddressPublisher{}{{IEEE}}.
  \newblock
  \begin{APACrefDOI} \doi{10.1109/ijcnn.1991.170605} \end{APACrefDOI}
  \PrintBackRefs{\CurrentBib}
  
  \bibitem [\protect \citeauthoryear {%
    Singh%
    , Barto%
    \BCBL {}\ \BBA {} Chentanez%
  }{%
    Singh%
    \ \protect \BOthers {.}}{%
    {\protect \APACyear {2004}}%
  }]{%
    SinghBartoEtAl-04}
  \APACinsertmetastar {%
    SinghBartoEtAl-04}%
  \begin{APACrefauthors}%
    Singh, S.%
    , Barto, A\BPBI G.%
    \BCBL {}\ \BBA {} Chentanez, N.%
  \end{APACrefauthors}%
  \unskip\
  \newblock
  \APACrefYearMonthDay{2004}{}{}.
  \newblock
  {\BBOQ}\APACrefatitle {Intrinsically Motivated Reinforcement Learning}
  {Intrinsically motivated reinforcement learning}.{\BBCQ}
  \newblock
  \BIn{} \APACrefbtitle {Proceedings of the 17th International Conference on
    Neural Information Processing Systems} {Proceedings of the 17th 
    international
    conference on neural information processing systems}\ (\BPG~1281–1288).
  \newblock
  \APACaddressPublisher{Cambridge, MA, USA}{MIT Press}.
  \PrintBackRefs{\CurrentBib}
  
  \bibitem [\protect \citeauthoryear {%
    Smithers%
  }{%
    Smithers%
  }{%
    {\protect \APACyear {1997}}%
  }]{%
    SMITHERS199788}
  \APACinsertmetastar {%
    SMITHERS199788}%
  \begin{APACrefauthors}%
    Smithers, T.%
  \end{APACrefauthors}%
  \unskip\
  \newblock
  \APACrefYearMonthDay{1997}{}{}.
  \newblock
  {\BBOQ}\APACrefatitle {{Autonomy in Robots and Other Agents}} {{Autonomy 
  in
      Robots and Other Agents}}.{\BBCQ}
  \newblock
  \APACjournalVolNumPages{Brain and Cognition}{34}{1}{88--106}.
  \newblock
  \begin{APACrefDOI} \doi{https://doi.org/10.1006/brcg.1997.0908}
  \end{APACrefDOI}
  \PrintBackRefs{\CurrentBib}
  
  \bibitem [\protect \citeauthoryear {%
    {Sphero, Inc.}%
  }{%
    {Sphero, Inc.}%
  }{%
    {\protect \APACyear {2020}}%
  }]{%
    sphero-bb8}
  \APACinsertmetastar {%
    sphero-bb8}%
  \begin{APACrefauthors}%
    {Sphero, Inc.}%
  \end{APACrefauthors}%
  \unskip\
  \newblock
  \APACrefYearMonthDay{2020}{}{}.
  \newblock
  \APACrefbtitle {{Licensed products: BB-8}.} {{Licensed products: BB-8}.}
  \newblock
  \APACrefnote{\url{https://support.sphero.com/category/kxwbdyqeyq-bb-8},
    accessed December 2020}
  \PrintBackRefs{\CurrentBib}
  
  \bibitem [\protect \citeauthoryear {%
    Steels%
  }{%
    Steels%
  }{%
    {\protect \APACyear {2004}}%
  }]{%
    Steels-04}
  \APACinsertmetastar {%
    Steels-04}%
  \begin{APACrefauthors}%
    Steels, L.%
  \end{APACrefauthors}%
  \unskip\
  \newblock
  \APACrefYearMonthDay{2004}{}{}.
  \newblock
  {\BBOQ}\APACrefatitle {{The Autotelic Principle}} {{The Autotelic
      Principle}}.{\BBCQ}
  \newblock
  \BIn{} \APACrefbtitle {{Embodied Artificial Intelligence}} {{Embodied
      Artificial Intelligence}}\ (\BVOL\ 3139, \BPGS\ 231--242).
  \newblock
  \APACaddressPublisher{}{Springer Berlin Heidelberg}.
  \newblock
  \begin{APACrefDOI} \doi{10.1007/978-3-540-27833-7_17} \end{APACrefDOI}
  \PrintBackRefs{\CurrentBib}
  
  \bibitem [\protect \citeauthoryear {%
    {Stubbs}%
    , {Hinds}%
    \BCBL {}\ \BBA {} {Wettergreen}%
  }{%
    {Stubbs}%
    \ \protect \BOthers {.}}{%
    {\protect \APACyear {2007}}%
  }]{%
    4136857}
  \APACinsertmetastar {%
    4136857}%
  \begin{APACrefauthors}%
    {Stubbs}, K.%
    , {Hinds}, P\BPBI J.%
    \BCBL {}\ \BBA {} {Wettergreen}, D.%
  \end{APACrefauthors}%
  \unskip\
  \newblock
  \APACrefYearMonthDay{2007}{}{}.
  \newblock
  {\BBOQ}\APACrefatitle {{Autonomy and Common Ground in Human-Robot 
  Interaction:
      A Field Study}} {{Autonomy and Common Ground in Human-Robot 
      Interaction: A
      Field Study}}.{\BBCQ}
  \newblock
  \APACjournalVolNumPages{IEEE Intelligent Systems}{22}{2}{42-50}.
  \newblock
  \begin{APACrefDOI} \doi{10.1109/MIS.2007.21} \end{APACrefDOI}
  \PrintBackRefs{\CurrentBib}
  
  \bibitem [\protect \citeauthoryear {%
    Thrun%
    , Burgard%
    \BCBL {}\ \BBA {} Fox%
  }{%
    Thrun%
    \ \protect \BOthers {.}}{%
    {\protect \APACyear {2005}}%
  }]{%
    ThrunBurgardEtAl-05}
  \APACinsertmetastar {%
    ThrunBurgardEtAl-05}%
  \begin{APACrefauthors}%
    Thrun, S.%
    , Burgard, W.%
    \BCBL {}\ \BBA {} Fox, D.%
  \end{APACrefauthors}%
  \unskip\
  \newblock
  \APACrefYear{2005}.
  \newblock
  \APACrefbtitle {{Probabilistic Robotics}} {{Probabilistic Robotics}}.
  \newblock
  \APACaddressPublisher{}{MIT Press}.
  \PrintBackRefs{\CurrentBib}
  
  \bibitem [\protect \citeauthoryear {%
    Tremoulet%
    \ \BBA {} Feldman%
  }{%
    Tremoulet%
    \ \BBA {} Feldman%
  }{%
    {\protect \APACyear {2000}}%
  }]{%
    TremouletFeldman-00}
  \APACinsertmetastar {%
    TremouletFeldman-00}%
  \begin{APACrefauthors}%
    Tremoulet, P\BPBI D.%
    \BCBT {}\ \BBA {} Feldman, J.%
  \end{APACrefauthors}%
  \unskip\
  \newblock
  \APACrefYearMonthDay{2000}{}{}.
  \newblock
  {\BBOQ}\APACrefatitle {{Perception of Animacy from the Motion of a Single
      Object}} {{Perception of Animacy from the Motion of a Single 
      Object}}.{\BBCQ}
  \newblock
  \APACjournalVolNumPages{Perception}{29}{8}{943--951}.
  \newblock
  \begin{APACrefDOI} \doi{10.1068/p3101} \end{APACrefDOI}
  \PrintBackRefs{\CurrentBib}
  
  \bibitem [\protect \citeauthoryear {%
    Van~Rensburg%
    , Rechnitzer%
    , Causo%
    \BCBL {}\ \BBA {} Whittington%
  }{%
    Van~Rensburg%
    \ \protect \BOthers {.}}{%
    {\protect \APACyear {2001}}%
  }]{%
    VanRensburgRechnitzerEtAl-01}
  \APACinsertmetastar {%
    VanRensburgRechnitzerEtAl-01}%
  \begin{APACrefauthors}%
    Van~Rensburg, E\BPBI J\BPBI J.%
    , Rechnitzer, A.%
    , Causo, M\BPBI S.%
    \BCBL {}\ \BBA {} Whittington, S\BPBI G.%
  \end{APACrefauthors}%
  \unskip\
  \newblock
  \APACrefYearMonthDay{2001}{}{}.
  \newblock
  {\BBOQ}\APACrefatitle {{Self-averaging sequences in the statistical 
  mechanics
      of random copolymers}} {{Self-averaging sequences in the statistical
      mechanics of random copolymers}}.{\BBCQ}
  \newblock
  \APACjournalVolNumPages{Journal of Physics A: Mathematical and
    General}{34}{33}{6381}.
  \newblock
  \begin{APACrefURL} \url{http://stacks.iop.org/ja/34/6381} \end{APACrefURL}
  \PrintBackRefs{\CurrentBib}
  
  \bibitem [\protect \citeauthoryear {%
    Wojciszke%
    , Bazinska%
    \BCBL {}\ \BBA {} Jaworski%
  }{%
    Wojciszke%
    \ \protect \BOthers {.}}{%
    {\protect \APACyear {1998}}%
  }]{%
    WojciszkeBazinskaEtAl-98}
  \APACinsertmetastar {%
    WojciszkeBazinskaEtAl-98}%
  \begin{APACrefauthors}%
    Wojciszke, B.%
    , Bazinska, R.%
    \BCBL {}\ \BBA {} Jaworski, M.%
  \end{APACrefauthors}%
  \unskip\
  \newblock
  \APACrefYearMonthDay{1998}{}{}.
  \newblock
  {\BBOQ}\APACrefatitle {{On the Dominance of Moral Categories in Impression
      Formation}} {{On the Dominance of Moral Categories in Impression
      Formation}}.{\BBCQ}
  \newblock
  \APACjournalVolNumPages{{Personality and Social Psychology
      Bulletin}}{24}{12}{1251--1263}.
  \newblock
  \begin{APACrefDOI} \doi{10.1177/01461672982412001} \end{APACrefDOI}
  \PrintBackRefs{\CurrentBib}
  
  \bibitem [\protect \citeauthoryear {%
    Yatani%
  }{%
    Yatani%
  }{%
    {\protect \APACyear {2016}}%
  }]{%
    Yatani-16}
  \APACinsertmetastar {%
    Yatani-16}%
  \begin{APACrefauthors}%
    Yatani, K.%
  \end{APACrefauthors}%
  \unskip\
  \newblock
  \APACrefYearMonthDay{2016}{}{}.
  \newblock
  {\BBOQ}\APACrefatitle {{Effect Sizes and Power Analysis in HCI}} {{Effect 
  Sizes
      and Power Analysis in HCI}}.{\BBCQ}
  \newblock
  \BIn{} J.~Robertson\ \BBA {} M.~Kaptein\ (\BEDS), \APACrefbtitle {{Modern
      Statistical Methods for HCI}} {{Modern Statistical Methods for HCI}}\ 
      (\BPGS\
  87--110).
  \newblock
  \APACaddressPublisher{}{Springer}.
  \newblock
  \begin{APACrefDOI} \doi{10.1007/978-3-319-26633-6_5} \end{APACrefDOI}
  \PrintBackRefs{\CurrentBib}
  
  \bibitem [\protect \citeauthoryear {%
    Zahedi%
    , Martius%
    \BCBL {}\ \BBA {} Ay%
  }{%
    Zahedi%
    \ \protect \BOthers {.}}{%
    {\protect \APACyear {2013}}%
  }]{%
    ZahediMartiusEtAl-13}
  \APACinsertmetastar {%
    ZahediMartiusEtAl-13}%
  \begin{APACrefauthors}%
    Zahedi, K.%
    , Martius, G.%
    \BCBL {}\ \BBA {} Ay, N.%
  \end{APACrefauthors}%
  \unskip\
  \newblock
  \APACrefYearMonthDay{2013}{}{}.
  \newblock
  {\BBOQ}\APACrefatitle {{Linear combination of one-step predictive 
  information
      with an external reward in an episodic policy gradient setting: a 
      critical
      analysis}} {{Linear combination of one-step predictive information 
      with an
      external reward in an episodic policy gradient setting: a critical
      analysis}}.{\BBCQ}
  \newblock
  \APACjournalVolNumPages{Frontiers in Psychology}{4}{}{}.
  \newblock
  \begin{APACrefDOI} \doi{10.3389/fpsyg.2013.00801} \end{APACrefDOI}
  \PrintBackRefs{\CurrentBib}
\end{thebibliography}
\end{document}